\documentclass[12pt]{iopart}

\usepackage{iopams}  
\usepackage{graphicx}
\usepackage{cite}
\usepackage{hyperref}
\usepackage{cleveref}
\usepackage{placeins}
\usepackage{float}
\usepackage[font=small]{caption}
\usepackage{longtable}
\usepackage{tabularx, booktabs}
\usepackage{pifont}
\usepackage{lscape}

\usepackage[switch]{lineno}

\newcounter{procstep}
\newcolumntype{N}{>{
    \addtocounter{procstep}{1}}c
}

\begin{document}

\title[Quantifying the biomimicry gap in biohybrid robot-fish pairs]{Quantifying the biomimicry gap in biohybrid robot-fish pairs}

\author{Vaios Papaspyros$^{1}$, Guy Theraulaz$^{2}$, Cl\'ement Sire$^{3}$, and Francesco Mondada$^{1}$}
 \vskip 0.2cm
\address{$^{1}$Mobile Robotic Systems (MOBOTS) group, School of Computer Science, \'Ecole Polytechnique Fédérale de Lausanne (EPFL), CH-1015 Lausanne, Switzerland. {\tt\small vaios.papaspyros@gmail.com, francesco.mondada@epfl.ch} \\  \vskip 0.2cm
$^{2}$Centre de Recherches sur la Cognition Animale, Centre de Biologie Int\'egrative, CNRS, Universit\'e de Toulouse III -- Paul Sabatier, 31062 Toulouse, France. {\tt\small guy.theraulaz@univ-tlse3.fr} \\ \vskip 0.2cm
$^{3}$Laboratoire de Physique Th\'eorique, CNRS, Universit\'e de Toulouse III -- Paul Sabatier, 31062 Toulouse, France. {\tt\small clement.sire@univ-tlse3.fr}}
\vspace{10pt}
\begin{indented}
    \item[] March 2024
\end{indented}

\begin{abstract}
    Biohybrid systems in which robotic lures interact with animals have become compelling tools for probing and identifying the mechanisms underlying collective animal behavior. One key challenge lies in the transfer of social interaction models from simulations to reality, using robotics to validate the modeling hypotheses. This challenge arises in bridging what we term the ``biomimicry gap'', which is caused by imperfect robotic replicas, communication cues and physics constraints not incorporated in the simulations, that may elicit unrealistic behavioral responses in animals. In this work, we used a biomimetic lure of a rummy-nose tetra fish (\textit{Hemigrammus rhodostomus}) and a neural network (NN) model for generating biomimetic social interactions. Through experiments with a biohybrid pair comprising a fish and the robotic lure, a pair of real fish, and simulations of pairs of fish, we demonstrate that our biohybrid system generates social interactions mirroring those of genuine fish pairs. Our analyses highlight that: 1) the lure and NN maintain minimal deviation in real-world interactions compared to simulations and fish-only experiments, 2) our NN controls the robot efficiently in real-time, and 3) a comprehensive validation is crucial to bridge the biomimicry gap, ensuring realistic biohybrid systems.
\end{abstract}

%
\vspace{2pc}
\vspace{2pc}
%
%
\maketitle
%
%

\begin{figure}[ht!]
    \centering
    \includegraphics[width=0.7\linewidth]{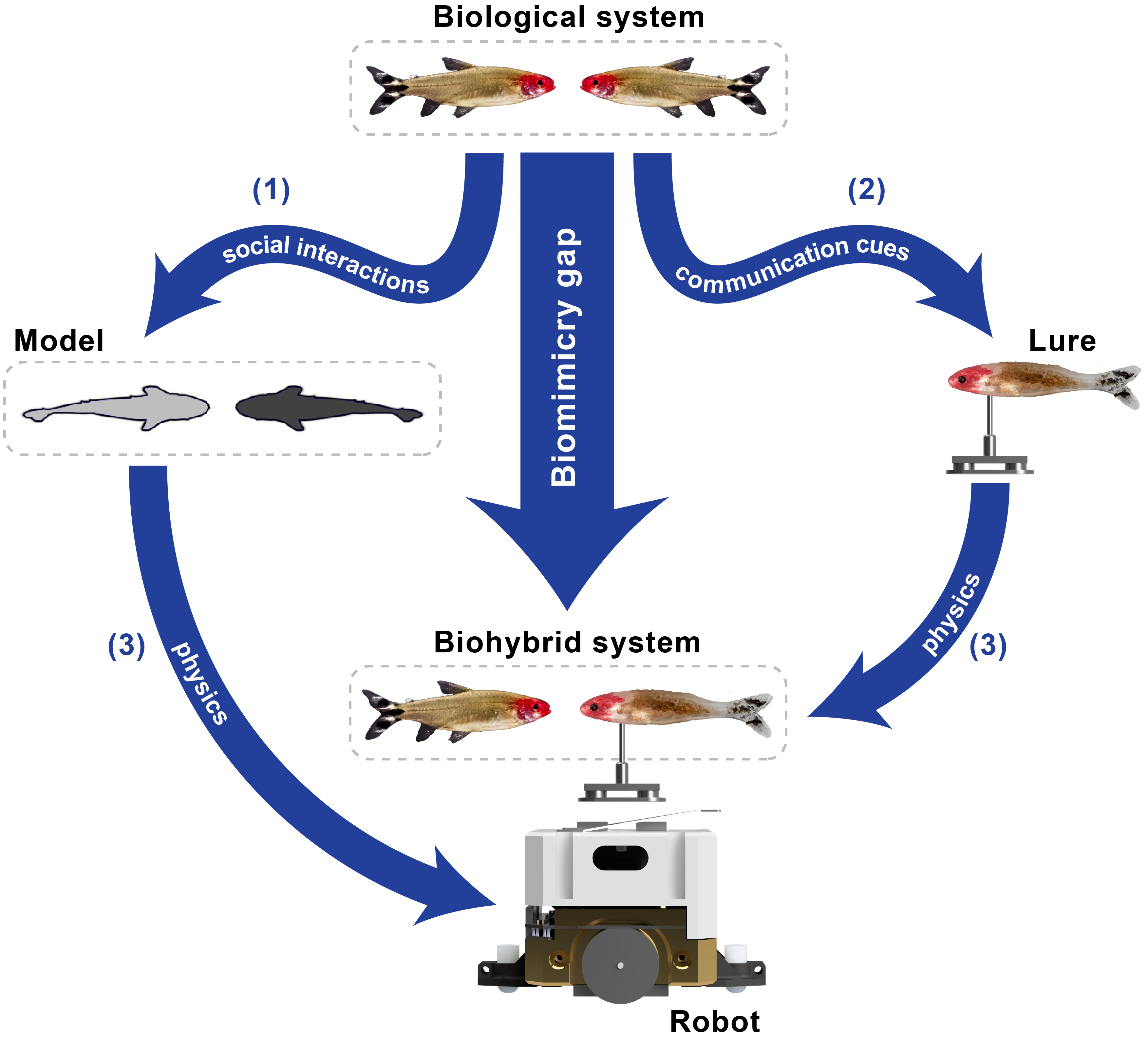}
    \caption{\small\textbf{Illustration of the sources of the biomimicry gap}. (1) The modeling phase may introduce a first source of discrepancy between the effect of social interactions on the swimming patterns in the model and the ones observed in real fish. (2) A second source of discrepancy between the visual appearance of the lure and that of a real fish might introduce imperfect communication cues and elicit unrealistic behavioral responses from neighboring organisms. (3) Finally, a third source of discrepancy between the characteristics of the movement produced by the model and its realization by the lure occurs when the numerical model is transferred to real-world scenarios due to the physics constrains that were not accounted for in the model. \textit{H.~rhodostomus} photo was taken by David Villa ScienceImage/CBI/CNRS, Toulouse.}
    \label{fig:biomimicry_gap}
\end{figure}

\begin{figure}[t!]
    \centering
    \includegraphics[width=\linewidth]{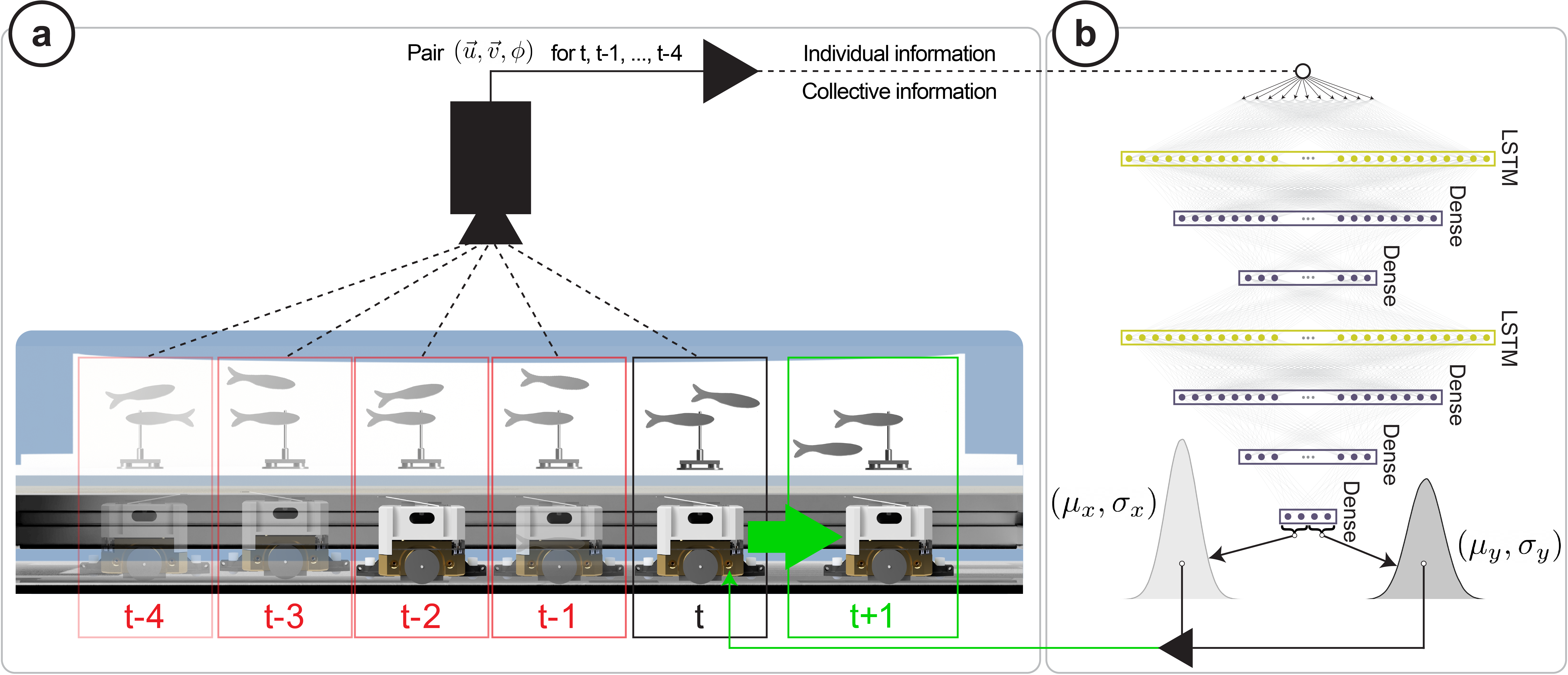}
    \caption{\small\textbf{Closed-loop robot control with Deep Learning Interaction (DLI) model}.
        \textbf{(a)} We use the top setup camera to track all agents (fish and/or lure) in real-time, and store unique trajectories for each agent. A $5 \times 11$ vector of individual and collective states, spanning 5 timesteps is fed to the DLI.
        \textbf{(b)} The DLI outputs two acceleration distributions, one for each Cartesian component. Then, the acceleration is used to compute the updated desired speed and position for $t+1$, which are communicated to the robot.
    }
    \label{fig:system}
\end{figure}


Robot-animal interactions have been increasingly gaining momentum as means to study collective behavior. Biohybrid systems, composed of living organisms and artificial agents, are particularly compelling as they enable researchers to investigate the way animals respond to controlled interactions. This is typically achieved through autonomous robotic devices equipped with species-specific communication channels, which can be employed to evoke responses in a biomimetic or non-biomimetic manner \cite{romano2019review}. Robots offer the advantage of conducting repetitive and repeatable experiments, even when driven by complex behavioral models. This is particularly important in the context of social interactions, which encompass considerable complexity when scaling from short-term interactions at the individual level to long-term emergent collective patterns.

Social fish species, such as the rummy-nose tetra (\textit{Hemigrammus rhodostomus}) and zebrafish (\textit{Danio rerio}), are frequently selected for these studies due to the intricacy of their short- and long-term interactions and their suitability for laboratory environments \cite{butail2015fish, romano2019review}, as well as the abundance of general knowledge about their behavior, genetics, and housing conditions \cite{orger2017zebrafish, miller2012schooling, zienkiewicz2018data, collignon2016stochastic, harpaz2021collective}. As a matter of fact, many fish-robot systems have been proposed to investigate various aspects of fish behavior, employing behavioral models with diverse degrees of detail and realistic features, and typically relying on analytical modeling approaches based on observation of fish interaction \cite{faria2010novel, swain2011real, landgraf2013interactive, bonnet2014miniature, landgraf2016robofish, bonnet2018closed, porfiri2018inferring, papaspyros2019bidirectional, romano2021unveiling, romano2022robot, cazenille2018mimetic, cazenille2018blend, cazenille2017automated, collignon2016stochastic}. Concurrently, machine learning-based modeling approaches have gained a growing interest \cite{papaspyros2024predicting, heras2018deep, costa2020automated, cazenille2018evolutionary, cazenille2019automatic}, but only a handful have been tested in real-time with a robotic device \cite{cazenille2018evolutionary}. These machine learning approaches are usually intended to study collective behavior by predicting motion in simulations alone \cite{heras2018deep, costa2020automated, cazenille2018evolutionary}, while the studies that exploit robotic systems typically evaluate only instantaneous group-level quantities in the short-term timescale \cite{cazenille2019automatic}.

A few flocking models for fish behavior, analytical or machine learning, have been evaluated in extended simulations to study long-term emergent collective behavior \cite{collignon2016stochastic, calovi2018disentangling, papaspyros2024predicting}. However, these models have not been tested and validated in biohybrid groups. Conversely, numerous models have been implemented on robotic devices without being tested in simulations \cite{abaid2010fish, bonnet2016infiltrating, cazenille2017automated, cazenille2018mimetic, cazenille2018blend, cazenille2019automatic, papaspyros2019bidirectional}. Furthermore, the majority of these studies involve robot experiments lasting no more than 30 minutes, with the resulting interaction patterns being rarely or only superficially quantified. Consequently, none of these models have been stringently benchmarked on both short- and long-term timescales within both simulation and fish-robot biohybrid experiments. In fact, previous research indicates that certain models may yield satisfactory biomimetic outcomes in the short term while failing to reproduce emergent dynamics accurately on longer time scales \cite{papaspyros2024predicting}.

Moreover, the transfer of computer models of social interactions into robot controllers that operate in real situations involving animals is not straightforward and can generate a discrepancy with the corresponding numerical simulations, akin to the reality gap observed when transferring simulated robot controllers to real-world applications \cite{mouret201720}.
As depicted in Fig.~\ref{fig:biomimicry_gap}, several sources of discrepancy can combine and feed this gap: 
\begin{enumerate}
    \item[(1)] subtle behavioral patterns that social interaction models may fail to capture; 
    
    \item[(2)] non-trivial physics related to the operation of the robot in real life that were not accounted for;
    \item[(3)] the extent of biomimicry exhibited by artificial lures \cite{romano2022any, papaspyros2023biohybrid}. 
\end{enumerate}
We refer to the cumulative effect of these discrepancies with the term ``biomimicry gap''.
Therefore, the biomimicry gap is an inherent aspect of the multifaceted, cross-domain process of creating biohybrid groups composed of animals and robots. To the best of our knowledge, the feasibility of bridging this biomimicry gap --- achieved by conducting extended experiments in both simulated and real-world environments, and comparing their results --- has yet to be conclusively and rigorously validated across all these levels in a single approach.

In this study, we investigate this notion by employing the (pretrained) machine learning model presented in \cite{papaspyros2024predicting}. We implement this model on a robotic system which is shown to achieve unprecedented levels of biomimicry, the LureBot \cite{papaspyros2023biohybrid}. 
The LureBot consists of an agile mobile robot capable of generating accelerations and velocities that closely mimic those of \textit{H.~rhodostomus}. The robot moves between two plates under the tank where the real fish swim. Additionally, a highly biomimetic artificial lure is magnetically attached to the robot and moves in the same tank as the fish. This lure is meticulously designed to faithfully resemble a real \textit{H.~rhodostomus}.
We exploit the robotic system to execute approximately 11 hours of multiple pair experiments wherein the biomimetic lure interacts with a single \textit{H.~rhodostomus}. This allows us to measure the behavioral differences between actual and simulated pairs \textit{H.~rhodostomus}, as well as, pairs of 1 biomimetic lure and 1 \textit{H.~rhodostomus}. In turn, this yields the first end-to-end approach aimed at minimizing the biomimicry gap, and presented in the following sections.

\section{Methods}

\subsection{Real-time tracking and robot control}
\label{sec:controlandtracking}

Experiments were performed with the Behavioral Observation and Biohybrid Interaction (BOBI) framework \cite{papaspyros2023biohybrid}, including the LureBot, to propel a \textit{H.~rhodostomus} lure. A $30$\,Hz camera mounted at the top of the setup keeps track of fish and the artificial lure swimming inside a circular water tank of radius $R = 25$\,cm. In addition,  a second $30$\,Hz camera on the bottom of the setup tracks the LureBot. The information is combined to distinguish which of the individuals seen by the top camera is the lure. Furthermore, BOBI is able to track multiple agents (here, only 2 are used) in real time, while maintaining unique IDs for each agent's trajectory. These agent-specific sequences of spatial movement can be exploited  by a behavioral model (see Section~\ref{sec:dli}) to compute real-time individual and collective quantities concerning the biohybrid group, and close the loop of interaction by adapting the robot's behavior with instructions on future movements.

In BOBI, the output of such a behavioral model is communicated to a motion controller and converted to motor commands for the differential drive of the LureBot. Here, we use the Proportional-Integral-Derivative (PID) controller as defined in BOBI, that incorporates a priori velocity information provided by the behavioral model. The PID combines the linear and angular errors between the LureBot's current and desired position, as well as the model's predicted velocity profile, to smoothly displace the robot.

\subsection{Deep Learning Interaction model}
\label{sec:dli}

We use a pretrained version of the Deep Learning Interaction (DLI) model \cite{papaspyros2024predicting}, to generate real-time goal positions for the LureBot \cite{papaspyros2023biohybrid}. The DLI consists of 7 layers (see Fig.~\ref{fig:system}b): 1st and 4th are LSTM layers \cite{hochreiter1997long}; the remaining are densely connected layers; ReLU activations are used for all layers except for the last one which is linear. For a single agent $i$, the state at time $t$ is defined as a $1 \times 5$ vector ${\bf s}_i(t)$:
\begin{equation}
    {\bf s}_{i}(t) = \big(\vec{u}_i(t), \vec{v}_i(t), r_{\rm w}^i(t) \big)
    \in \mathbb{R}^5,
    \label{eq:indstate}
\end{equation}
where $\vec{u}_i(t)$, $\vec{v}_i(t)$ are the 2D position and velocity, respectively, and where $r_{\rm w}^i(t)$ is the distance of the individual $i$ from the wall at time $t$. Then, the pairwise state at time $t$ is summarized in the following $1 \times 11$ vector:
\begin{equation}
    {\bf S}_{ij}(t) = \big(\underbrace{{\bf s}_{i}(t)}_{\textrm{\tiny individual (focal)}\atop\textrm{\tiny information}}, \underbrace{{\bf s}_{j}(t)}_{\textrm{\tiny individual (neighbor)}\atop\textrm{\tiny information}},~\underbrace{d_{ij}(t)}_{\textrm{\tiny collective}\atop\textrm{\tiny information}} ~\big)
    \in \mathbb{R}^{11},
    \label{eq:input}
\end{equation}
with $i$ the focal individual for which we generate trajectory predictions, $j$ its neighbor, and $d_{ij}$ their interindividual distance. In real-time, we feed the DLI with a $5 \times 11$ sequence $({\bf S}(t-4), \dots, {\bf S}(t))$ of the pair-wise states (see Fig.~\ref{fig:system}a), where we make sure that $i$ (focal individual) corresponds to the LureBot.

Subsequently, the DLI model outputs the expected acceleration mean and standard deviation value, $(\mu_x,\sigma_x)$ and $(\mu_y,\sigma_y)$, of the Cartesian components $x$ and $y$. Assuming a Gaussian distribution for the acceleration~\cite{chua2018deep},  we sample this distribution to produce acceleration predictions $\vec{a} = (a_x,a_y)$ and use the following motion equations to generate velocity commands and the goal position of the LureBot at time $t+1$:
\begin{eqnarray}
    \vec{v}_i(t+1) &=& \vec{v}_i(t) + \Delta t \, \vec{a},\label{eq:motion_eqs1}\\
    \vec{u}_i(t+1) &=& \vec{u}_i(t) + \Delta t \, \vec{v}_i(t+1),
    \label{eq:motion_eqs2}
\end{eqnarray}
where $\Delta t = 0.12\,s$, a choice made with respect to the data filtering procedure applied on the raw data to generate an intermediate training dataset for the DLI \cite{papaspyros2024predicting}.
The 2D velocity commands, defined in equation~\ref{eq:motion_eqs1}, and goal position, defined in equation~\ref{eq:motion_eqs2}, are given to the BOBI's PID \cite{papaspyros2023biohybrid}, and eventually translated to motor commands (see Section~\ref{sec:controlandtracking}; see Fig.~\ref{fig:system}a,b).

In \cite{papaspyros2024predicting}, this approach  was validated in long simulations and was shown to be capable of reproducing the social dynamics of \textit{H.~rhodostomus} pairs faithfully compared to experiments. In the following sections, we test the extent to which the DLI can produce faithful interactions when deployed on a physical robot-fish group instead of a simulated group.

\begin{figure}[t!]
    \centering
    \includegraphics[width=0.7\linewidth]{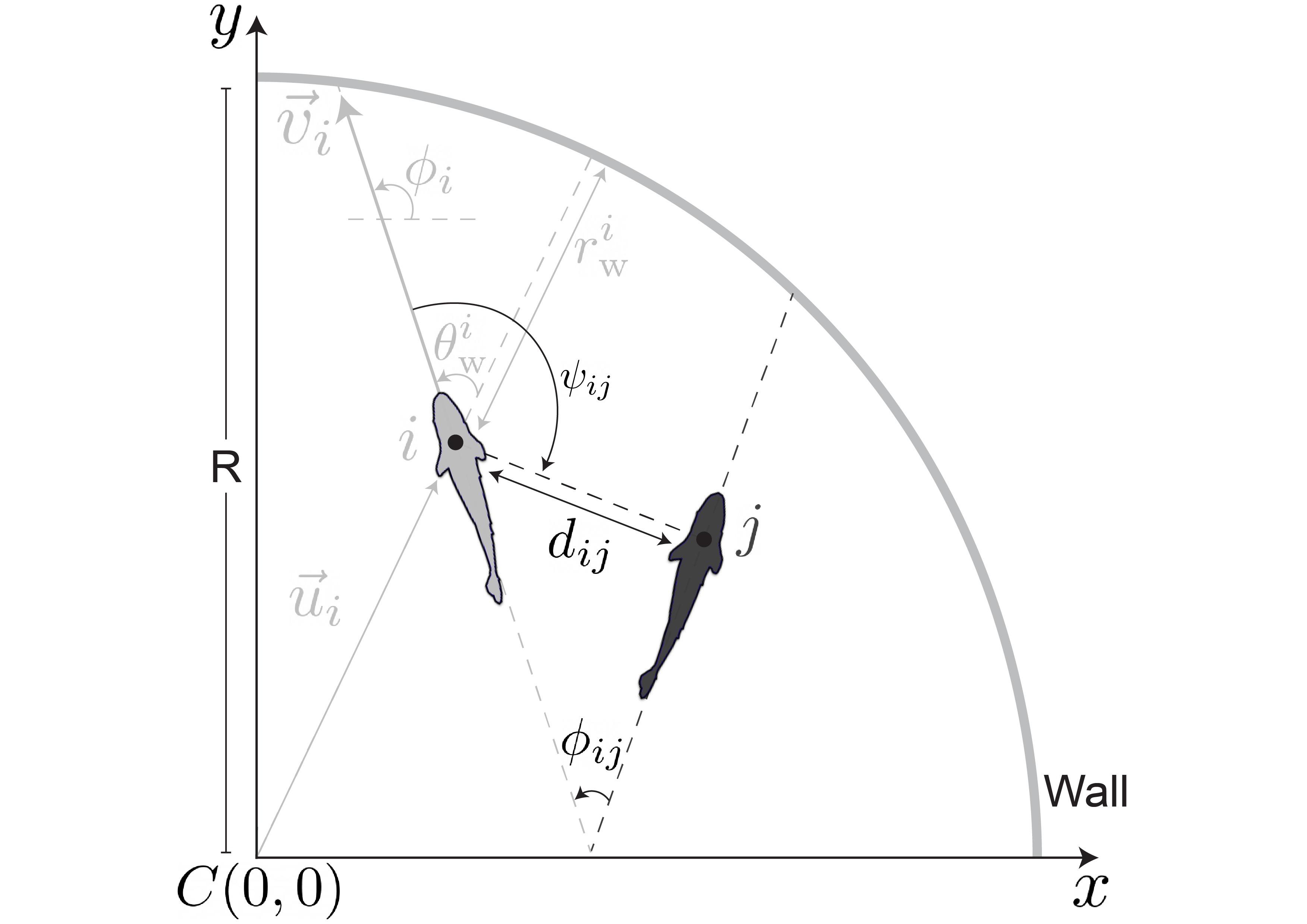}
    \caption{\small\textbf{Individual and collective variables}. For the focal individual $i$ (light gray), we define the individual quantities: $\vec{u}_i$, its Cartesian position; $\vec{v}_i$, its instantaneous velocity; $r_{\rm w}^i$, its distance to the wall;  $\phi_{i}$, its heading angle. We also define the collective quantities from $i$'s perspective when another individual $j$ (dark gray) is also present in the circular tank of radius $R=25$\,cm: $d_{ij}$, the interindividual distance; $\phi_{ij}$, the heading difference between both individuals; $\psi_{ij}$, the angle with which individual $j$ is perceived by the focal individual $i$. Note that, for visualization purposes, the size of agents is not to scale.}
    \label{fig:quantities}
\end{figure}

\subsection{Evaluating the outcome of short- and long-term interactions between fish and the LureBot}
\label{sec:metrics}

Evaluating the extent to which models can reproduce the social dynamics of animal groups, here \textit{H.~rhodostomus} fish, is a non-trivial task. As explored in \cite{papaspyros2024predicting}, such models may succeed in reproducing quantities in the short-term timescale, but may also fail to reproduce the emergent dynamics in the long term. Here, we opt to benchmark our results by exploiting the 9 observables considered in \cite{papaspyros2024predicting, jayles2020collective} (see Fig.~\ref{fig:quantities}).

The first 3 observables correspond to instantaneous quantities at the \textit{individual} level, for which we measure their probability density function (PDF): the speed, $V$ of an individual; its distance to the wall, $r_{\rm w}$; and its heading angle relative to the normal to the wall, $\theta_{\rm w}$. 3 additional observables probe the instantaneous \textit{collective} dynamics: the  distance $d_{ij}$ between the pair of individuals; the difference $|\phi_{ij}|$ between the heading directions of the two individuals; and the angle $\psi_{ij}$ at which an individual perceives its neighbor. Finally, we consider 3 \textit{temporal correlation functions} that probe the social dynamics at a very fine level~\cite{jayles2020collective}, and which are generally particularly difficult to reproduce:


\begin{equation}
    C_X(t) = \left\langle\left[\vec{u}_i(t + t') - \vec{u}_i(t')\right]^ 2\right\rangle,
    \label{eq:cx}
\end{equation}

\begin{equation}
    C_V(t) = \left\langle \vec{v_i}(t + t') \cdot \vec{v}_i(t')\right \rangle,
    \label{eq:cv}
\end{equation}

\begin{equation}
    C_{\theta_{\rm w}}(t) = \left\langle\cos\left[\theta_{\rm w}^i(t + t') - \theta_{\rm w}^i(t')\right]\right \rangle.
    \label{eq:ctheta}
\end{equation}

$C_X$ is the mean-squared displacement, $C_V$ the velocity autocorrelation, and $C_{\theta_{\rm w}}$ the autocorrelation of the angle of incidence to the wall. In general, we denote $C_q(t) = \langle q(t + t') q(t')  \rangle$ as the average of the quantity $q(t')q(t + t')$ over the reference times $t'$, over individuals, and over different experiments. Assuming the stationarity of the system, the temporal correlation function $C_q(t)$ only depends on the time difference between observations, and is often noted $C_q(t) = \langle q(t)q(0) \rangle$ (implicitly implying an average over the reference time $t' = 0$).

\subsection{Quantifying the  (dis)similarity between two PDF: the Hellinger distance}
\label{sec:Hellinger}
The comparison between the different test cases exploits the 9 observables introduced above and supplementary videos for fish-only experiments, DLI simulated pairs (DLI-SP), and biohybrid pairs (DLI-SP)\footnote{The videos are also available at \url{https://doi.org/10.5281/zenodo.8253256}}. 

For all quantities (PDF and correlation functions), we have computed the statistical and sample to sample standard error by using a bootstrap method. In addition, for each PDF, we report the mean and standard deviation (SD) in Table~\ref{tab:means_sds}, as well as their standard error that we will omit to mention in the hereafter analysis of the results, for readability (except when their value is relevant to the discussion). 

Moreover, in order to compare the PDF for a given quantity between two given test cases, we compute the Hellinger distance between these distributions in Table~\ref{tab:hellinger}. For two PDF $F$ and $G$ for the same quantity $x$, the  Hellinger distance $H(F|G)$ quantifies their (dis)similarity~\cite{basu19972, beran1977minimum}:
\begin{eqnarray}
    {H(F|G)} &=& \frac{1}{2}\int {{{\left( {\sqrt {F(x)}  - \sqrt {G(x)} } \right)}^2}} dx,\\
    &=& 1 - \int \sqrt {F(x)} \sqrt {G(x)}  \,dx,
    \label{eq:hellinger}
\end{eqnarray}
where we have used the normalization of the PDF, $\int F(x)\,dx=\int G(x)\,dx=1$, to obtain the last
equality.
The first definition of $H(F|G)$ clarifies that it measures the overall difference between
$F(x)$ and $G(x)$. Meanwhile, the second equivalent definition provides a comprehensive interpretation in terms of
the \textit{overlap} of both PDF. Indeed, the second definition measures the distance from unity of
the scalar product of  $\sqrt{F(x)}$ and $\sqrt{G(x)}$ seen as vectors of unit Euclidean norm (a
consequence of the normalization, $\int\sqrt{F(x)}^2\,dx= 1$). The  Hellinger distance is 0 if both PDFs are identical, and is bounded by 1, a limit reached if the distributions have a non-overlapping support. In general, a Hellinger distance $H(F|G)\lesssim 0.1$ points to a good agreement between both PDF, $0.1\lesssim H(F|G)\lesssim 0.2$ points to a fair similarity between them, while $H(F|G)\gtrsim 0.2$ indicates that the two distributions are significantly dissimilar.

\subsection{Data for the dynamics of pairs of agents}

In this work, we focus on the social dynamics that arise from pairwise interactions in three different conditions. First, we consider $\approx 11$\,h of experiments involving pairs of \textit{H.~rhodostomus}, to characterize and quantify the spontaneous social interactions when no artificial devices are present in the tank.

Secondly, we consider $\approx 16$\,h of effective trajectories for DLI simulated pairs (DLI-SP)~\cite{papaspyros2024predicting}, as a baseline to the robot's underlying model in ideal conditions. This DLI model was originally trained in  \cite{papaspyros2024predicting} on a different series of experimental data obtained in \cite{calovi2018disentangling} for the same species (\textit{H.~rhodostomus}), but in different conditions. More specifically, we used a different tank of the same radius $R=25$\,cm, but made of a higher-quality material that is compatible with our robotic system \cite{papaspyros2023biohybrid}. In addition, our lighting conditions (which greatly impact the fish behavior) are also slightly different and adapted to the constraints for the real-time fish tracking algorithm. We will also mention the results obtained after retraining the DLI model with the fish data considered in the present work, which we will refer to as the DLIv2-SP (see Table~\ref{tab:means_sds} and Figs.~\ref{fig:si-individual}-\ref{fig:si-correlation}).

Finally, we have conducted $\approx 11$\,h of experiments where the LureBot propels a biomimetic lure moving inside the circular arena, which is interacting in closed-loop with an actual \textit{H.~rhodostomus}. For brevity, in the following analysis of the results, we will simply refer to the LureBot and the lure attached to it as the LureBot. The LureBot is given a pre-trained copy of the DLI model of \cite{papaspyros2024predicting}, which is queried in real time to generate biomimetic trajectories (see Section~\ref{sec:dli}). We refer to these data as DLI biohybrid pairs (DLI-BP). We did not perform experiments with the LureBot trained with the DLIv2 model, since our experimental campaign obviously predated the training of the DLIv2 model, which required these new experimental results.

In all experiments involving fish pairs or LureBot-fish pairs, we have  explicitly designed a protocol which did not allow the use of the same fish in an experiment for at least $48$\,h after their first test, to avoid potential learning effects when the fish interact with the lure. The fish housing conditions and experiments have been approved by the local ethical committee (see Section~\ref{sec:ethics}) and are described in detail in \cite{papaspyros2023biohybrid}.

\section{Results}

This section reports the detailed comparison between the three test cases: (fish-only) experiments with pairs of \textit{H.~rhodostomus}; DLI simulated pairs (DLI-SP); DLI biohybrid pairs (DLI-BP), that consist of the LureBot interacting in closed loop with a \textit{H.~rhodostomus}. Our results are also qualitatively illustrated by a supplementary video (see \url{https://zenodo.org/doi/10.5281/zenodo.8253256}) displaying side-by-side trajectories for the three test cases.
In addition, at the end of this section, we will briefly present results for DLIv2 simulated pairs (DLIv2-SP; trained on the fish-only experimental data of the present work).

As mentioned above, this section will also exploit the results of Table~\ref{tab:means_sds} (means and SD and their standard error) and Table~\ref{tab:hellinger} (Hellinger distances between PDF corresponding to two different conditions). At the end of Table~\ref{tab:hellinger}, we also report the Hellinger distances resulting from the inherent variability observed in fish-only experiments. They were obtained by a bootstrap method by randomly splitting the 14 fish-only experiments in two sets. Then, for each pair of sets, we compute the corresponding Hellinger distances between their associated PDF and average the results over the random draws. We find a mean Hellinger distance (averaged over the 6 observables) of $\bar{H}=0.10$, which constitutes a baseline for comparing the results of fish-only experiments to other conditions involving biohybrid or simulated pairs.

\subsection{Instantaneous individual observables}\label{sec:indquant}

\begin{figure}[thpb]
    \centering
    \includegraphics[width=\linewidth]{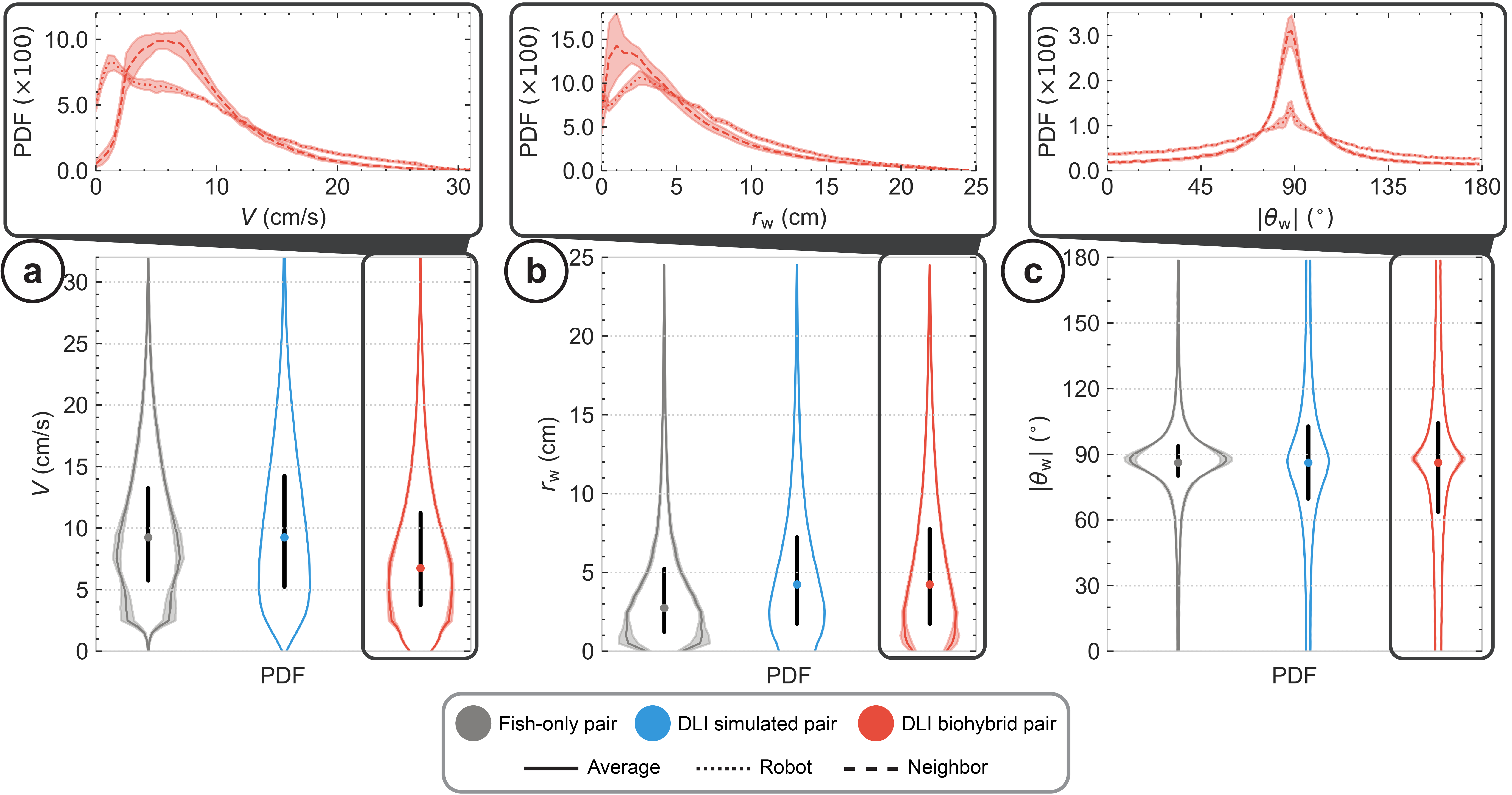}
    \caption{\small
        \textbf{Instantaneous individual quantities.} \textbf{(a)} Speed $V$ probability density function. \textbf{(b)} Distance to the wall $r_{\rm w}$ probability density function. \textbf{(c)} Angle of incidence to the wall $\theta_{\rm w}$ probability density function. Dark gray, blue, and red colors correspond to the distributions of the fish-only experiment, the DLI simulated pairs, and the DLI biohybrid pairs, respectively. In all PDFs, the colored dot corresponds to the median, and the thick horizontal black line corresponds to the limits of the first and third quartile. The top inset plots depict the PDFs of the DLI biohybrid pair experiments, where the dotted and dashed lines correspond to the robot's and its neighbor's distributions, respectively. The shaded areas correspond to the standard deviation.
    }
    \label{fig:individual}
\end{figure}

\begin{figure}[ht!]
    \centering
    \includegraphics[width=\linewidth]{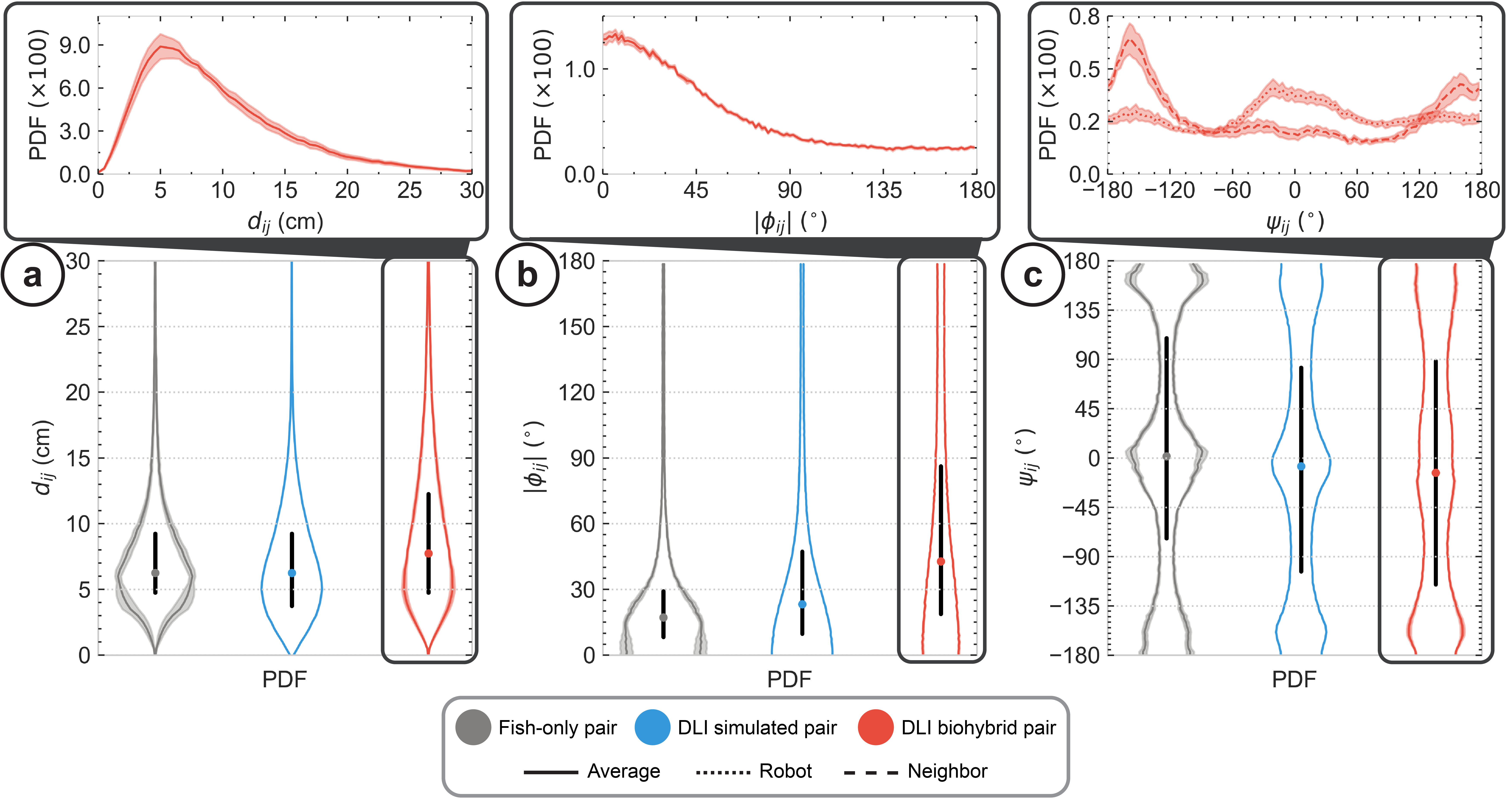}
    \caption{\small
        \textbf{Instantaneous collective quantities.} \textbf{(a)} Interindividual distance $d_{ij}$ probability density function. \textbf{(b)} Difference in heading angles $|\phi_{ij}|$ probability density function. \textbf{(c)} Viewing angle $\psi_{ij}$ probability density function. Dark gray, blue, and red colors correspond to the distributions of the experiment, DLI simulated pairs and DLI biohybrid pairs, respectively. In all PDFs, the colored dot corresponds to the median, and the thick horizontal black line corresponds to the limits of the first and third quartile. The inset plots depict the PDFs of the DLI biohybrid pair experiments where the dotted, dashed, and solid lines correspond to the robot, neighbor and average agent distributions, respectively. The shaded areas correspond to the standard deviation.
    }
    \label{fig:collective}
\end{figure}

Fig.~\ref{fig:individual}a shows the speed PDF  for  the three cases we considered. Fish pairs swim at a mean speed of $10.5$\,cm/s, associated to a standard deviation (SD) of 5.7\,cm/s (see Table~\ref{tab:means_sds}). DLI simulated pairs produce a rather similar speed PDF (Hellinger distance $H=0.09$; see Table~\ref{tab:hellinger}), albeit slightly wider (SD of 7.0\,cm/s), with a nearly identical mean of 11.1\,cm/s. For biohybrid pairs, the fish and the LureBot have a very similar mean speed (identical within error bars; see Table~\ref{tab:means_sds}), but which is 20\,\% smaller than in the fish-only experiments, although the SD is similar to that of the fish experiments, resulting in a Hellinger distance of $H=0.18$ (see Section~\ref{sec:Hellinger}).

In Fig.~\ref{fig:individual}b, we plot the PDF of  the distance to the wall, $r_w$, for each case. Fish pairs swim very close to the wall, with a mean distance of 4.4\,cm and a SD of 3.9\,cm, both comparable to the typical fish body length ($\sim$3.5\,cm). This is a consequence of the burst-and-coast swimming mode  exhibited by \textit{H.~rhodostomus}, as shown in \cite{calovi2018disentangling}. Indeed, the motion of this species is characterized by a succession of sudden acceleration periods (``kicks'' or bursts of typical duration 0.1\,s), each followed by a longer gliding period of typical duration 0.5\,s, during which the fish moves in a quasi straight line. Because of the rather narrow distribution of heading changes between kicks, even observed when a fish is far from the wall \cite{calovi2018disentangling}, the fish is unable to escape the concave boundaries of the wall, except when rare large heading changes occur.
The mean distance to the wall is 5.7\,cm for the DLI simulated pairs, and the associated PDF compared to fish experiments has a Hellinger distance of $H=0.13$, showing that the DLI model captures reasonably well the tendency of the fish to move close to the wall. For biohybrid pairs, we found that the fish swims farther from the wall than in fish-only experiments, with a mean distance of 5.5\,cm. In this case, the LureBot is even farther to the wall, at a mean distance of 6.6\,cm, which likely also causes the fish to swim farther to the wall than in fish-only experiments.

Finally, in Fig.~\ref{fig:individual}c, we plot the PDF of the absolute value of the heading angle relative to the normal to the wall, $|\theta_{\rm w}|$. As a consequence of the agents (fish, DLI model, or LureBot) moving close to the wall, we naturally find that the mean of $|\theta_{\rm w}|$ is very close to, but slightly below 90$^\circ$ (see Table~\ref{tab:means_sds}), with a difference which is statistically significant. Indeed, as already reported in the experiments of~\cite{calovi2018disentangling}, the agents spend slightly more time heading toward the wall ($|\theta_{\rm w}|<90^\circ$) than moving away from it ($|\theta_{\rm w}|>90^\circ$). The PDF for the three considered cases are symmetric around their mean, but we find that the fish experiments lead to the narrowest distribution, with a SD of 22$^\circ$, compared to a SD of 35$^\circ$ for the DLI simulated pairs, and a SD of 33$^\circ$ and 43$^\circ$ for the fish and the LureBot in a biohybrid pair. The values of these SD are naturally correlated with the mean distance of the agent to the wall: the farther the agent, the larger are the fluctuations (SD) of its heading angle relative to the wall. 

In summary, the DLI-BP system presents a fair agreement with the experimental results for all quantities. Concurrently, DLI-BP and DLI-SP show smaller dissimilarity, indicating that the transposition of the simulated model into the robot was successful (see Table~\ref{tab:hellinger}). We also observe that, in some cases (\textit{e.g.}, as reflected in the PDF of $|\theta_{\rm w}|$), the fish's behavior guides the DLI-powered robot, the latter  behaving more closely as a fish than the virtual/simulated DLI agents.
Nonetheless, the observables test the DLI's performance at a very fine level, especially in the case of DLI-BP, where the physical aspect is also impeding the precise reproduction of the social dynamics, either due to the imperfect (with respect to fish) motion of the robot or the varying degree the robotic system's acceptance by the fish.

\begin{table}
    \centering
    \resizebox{.65\textwidth}{!}{%
    \begin{tabular}{l|crr}
    \textbf{Pair}           & \textbf{Quantity}      & \textbf{Mean}    & \textbf{Standard deviation} \\
    \hline
                   &                        &                                                \\
    \textbf{Fish-only}      & $V$                    & $10.50 \pm 0.60$ & $5.73 \pm 0.36$             \\
                   & $r_{\textrm w}$        & $4.39 \pm 0.43$  & $3.86 \pm 0.22$             \\
                   & $|\theta_{\textrm w}|$ & $87.42 \pm 0.39$ & $21.91 \pm 1.46$            \\
                   & $d_{ij}$               & $8.05 \pm 0.71$  & $5.11 \pm 0.43$             \\
                   & $|\phi_{ij}|$          & $26.72 \pm 1.91$ & $29.81 \pm 1.24$            \\
                   & $\psi_{ij}$            & $7.96 \pm 4.73$  & $108.98 \pm 1.19$           \\
                   &                        &                                                \\
    \hline
                   &                        &                                                \\
    \textbf{DLI-SP }        & $V$                    & $11.06 \pm 0.04$ & $7.04 \pm 0.02$             \\
                   & $r_{\textrm w}$        & $5.66 \pm 0.03$  & $4.42 \pm 0.03$             \\
                   & $|\theta_{\textrm w}|$ & $88.07 \pm 0.06$ & $34.55 \pm 0.16$            \\
                   & $d_{ij}$               & $7.43 \pm 0.03$  & $4.38 \pm 0.04$             \\
                   & $|\phi_{ij}|$          & $38.06 \pm 0.19$ & $38.63 \pm 0.17$            \\
                   & $\psi_{ij}$            & $-4.11 \pm 0.33$ & $107.13 \pm 0.06$           \\
                   &                        &                                                \\
    \hline
                   &                        &                                                \\
    \textbf{DLI-BP}         & $V$                    & $8.60 \pm 0.22$  & $5.93 \pm 0.12$             \\
                   & $r_{\textrm w}$        & $6.05 \pm 0.25$  & $4.76 \pm 0.06$             \\
                   & $|\theta_{\textrm w}|$ & $86.44 \pm 0.17$ & $38.07 \pm 0.73$            \\
                   & $d_{ij}$               & $9.96 \pm 0.48$  & $6.27 \pm 0.33$             \\
                   & $|\phi_{ij}|$          & $58.60 \pm 0.91$ & $48.38 \pm 0.24$            \\
                   & $\psi_{ij}$            & $-7.42 \pm 4.16$ & $110.41 \pm 0.51$           \\
                   &                        &                  &                             \\
    \textbf{DLI-BP (fish)}  & $V$                    & $8.44 \pm 0.26$  & $5.13 \pm 0.21$             \\
                   & $r_{\textrm w}$        & $5.54 \pm 0.35$  & $4.54 \pm 0.09$             \\
                   & $|\theta_{\textrm w}|$ & $87.46 \pm 0.19$ & $32.76 \pm 1.25$            \\
                   &                        &                  &                             \\
    \textbf{DLI-BP (robot)} & $V$                    & $8.74 \pm 0.16$  & $6.62 \pm 0.12$             \\
                   & $r_{\textrm w}$        & $6.59 \pm 0.15$  & $4.91 \pm 0.05$             \\
                   & $|\theta_{\textrm w}|$ & $85.42 \pm 0.24$ & $42.78 \pm 0.79$            \\
                   &                        &                                                \\
    \hline
    \hline
                   &                        &                                                \\
    \textbf{DLIv2-SP }      & $V$                    & $10.53 \pm 0.48$ & $6.18 \pm 0.28$             \\
                   & $r_{\textrm w}$        & $4.64 \pm 0.23$  & $4.37 \pm 0.05$             \\
                   & $|\theta_{\textrm w}|$ & $87.56 \pm 0.11$ & $26.47 \pm 0.47$            \\
                   & $d_{ij}$               & $8.39 \pm 0.07$  & $6.15 \pm 0.11$             \\
                   & $|\phi_{ij}|$          & $30.54 \pm 0.30$ & $33.11 \pm 0.29$            \\
                   & $\psi_{ij}$            & $11.72 \pm 0.87$ & $109.08 \pm 0.19$           \\
   \end{tabular}%
   }     
   \caption{\small\textbf{Means and standard deviations.} For fish-only experiments, DLI simulated pairs (DLI-SP), and biohybrid pairs (DLI-SP), we report the mean and the standard deviation (SD) of the 6 observables introduced in Section~\ref{sec:metrics}, along with their respective standard error. The speed $V$ is given in cm/s, the distances $r_{\textrm w}$ and $d_{ij}$ are given in cm, and the angles $|\theta_{\textrm w}|$, $|\phi_{ij}|$, and $\psi_{ij}$ are in degrees. Note the small standard error in the case of the (DLI-SP) resulting from extensive simulations (16.6\,h long, almost twice the amount of data collected for other cases) and the fact that the 2 agents are statistically identical. For the biohybrid experiments, we report the mean and SD for $V$, $r_{\textrm w}$, and $|\theta_{\textrm w}|$, averaged over the fish and the LureBot, as well as for each of them. Finally, we present the corresponding results for a DLI model retrained on the present fish experiments (DLIv2-SP).} \label{tab:means_sds} 
\end{table}

\FloatBarrier
\begin{table}
\centering
    \resizebox{.66\textwidth}{!}{%
    \begin{tabular}{l|cc}
    \textbf{Pair}                   & \textbf{Quantity}      & \textbf{Hellinger distance $H$} \\
    \hline
                                    &                        &                             \\
    \textbf{Fish-only \textit{vs} DLI-SP}    & $V$           & $0.09$                      \\
                                    & $r_{\textrm w}$        & $0.13$                      \\
                                    & $|\theta_{\textrm w}|$ & $0.23$                      \\
                                    & $d_{ij}$               & $0.12$                      \\
                                    & $|\phi_{ij}|$          & $0.14$                      \\
                                    & $\psi_{ij}$            & $0.09$                      \\
                                    & Average $\bar{H}$      & $0.13$                        \\
                                    &                        &                             \\
    \hline
                                    &                        &                             \\
    \textbf{Fish-only \textit{vs} DLI-BP}    & $V$           & $0.18$                      \\
                                    & $r_{\textrm w}$        & $0.15$                      \\
                                    & $|\theta_{\textrm w}|$ & $0.25$                      \\
                                    & $d_{ij}$               & $0.16$                      \\
                                    & $|\phi_{ij}|$          & $0.30$                      \\
                                    & $\psi_{ij}$            & $0.15$                      \\
                                    & Average $\bar{H}$      & $0.20$                      \\
                                    &                        &                             \\
    \hline
                                    &                        &                             \\
    \textbf{DLI-SP \textit{vs} DLI-BP}       & $V$           & $0.14$                      \\
                                    & $r_{\textrm w}$        & $0.04$                      \\
                                    & $|\theta_{\textrm w}|$ & $0.04$                      \\
                                    & $d_{ij}$               & $0.18$                      \\
                                    & $|\phi_{ij}|$          & $0.17$                      \\
                                    & $\psi_{ij}$            & $0.07$                      \\
                                    & Average $\bar{H}$      & $0.11$                      \\
                                    &                        &                             \\
    \hline
    \hline
                                    &                        &                             \\
    \textbf{Fish-only \textit{vs}  DLIv2-SP} & $V$           & $0.05$                      \\
                                    & $r_{\textrm w}$        & $0.08$                      \\
                                    & $|\theta_{\textrm w}|$ & $0.08$                      \\
                                    & $d_{ij}$               & $0.14$                      \\
                                    & $|\phi_{ij}|$          & $0.06$                      \\
                                    & $\psi_{ij}$            & $0.04$                      \\
                                    & Average $\bar{H}$      & $0.08$                       \\
                                    &                        &                             \\

\hline
    \hline
                                    &                        &                             \\
    \textbf{Fish-only \textit{vs}  Fish-only (Bootstrap)} & $V$           & $0.07$                      \\
                                    & $r_{\textrm w}$        & $0.09$                      \\
                                    & $|\theta_{\textrm w}|$ & $0.04$                      \\
                                    & $d_{ij}$               & $0.16$                      \\
                                    & $|\phi_{ij}|$          & $0.15$                      \\
                                    & $\psi_{ij}$            & $0.09$                      \\
                                    & Average $\bar{H}$      & $0.10$                       \\
                                    &                        &                             \\                                    
\end{tabular}%
   }  
\caption{\small\rm\textbf{Hellinger distances.} We exploit the Hellinger distance between two PDF (see Section~\ref{sec:Hellinger}) to compare the PDF of the  6 observables introduced in Section~\ref{sec:metrics},  for fish-only experiments, DLI simulated pairs (DLI-SP and DLIv2-SP), and biohybrid pairs (DLI-BP). The last condition describes the inherent variability between the 14 fish experiments and is obtained by a bootstrap method by randomly splitting these 14 experiments in two sets, and computing the Hellinger distance between their 6 corresponding PDF. We also report the average Hellinger distance $\bar{H}$ for each condition.}
    \label{tab:hellinger}                                   
\end{table}

\subsection{Instantaneous collective observables}

\emph{H.~rhodostomus} have a natural tendency to swim in close proximity to each other. In our experiments, fish pairs typically maintain a median interindividual distance $d_{ij}$ of less than two body lengths (see Fig.~\ref{fig:collective}a), with a mean distance of $8.05\pm 0.7$\,cm and a SD of 5.1\,cm (see Table~\ref{tab:means_sds}). The dynamics of DLI simulated pairs results in a very similar PDF ($H =0.16$), with a mean of $7.43\pm 0.03$\,cm, which is within one standard error (for the fish experiments) from the mean obtained for fish. As for the biohybrid pair, it is less bound than pairs of fish or DLI, with a mean distance between the fish and the LureBot of $9.96\pm 0.5$\,cm. The distribution is also slightly wider, with a SD of 6.3\,cm. In fact, although the peak of the interindividual distance PDF is located at a similar value as for fish or DLI pairs ($5-6$\,cm in the three cases), the biohybrid pairs are more often separated by a distance larger than 15\,cm.

\emph{H.~rhodostomus} is a social species, often found to form well aligned schools. In fact, their pairwise alignment interaction was quantitatively measured in~\cite{calovi2018disentangling}, showing that this interaction remains strong up to three body lengths, well within the typical distance between fish. In Fig.~\ref{fig:collective}b, to quantify the alignment within pairs of agents, we plot the distribution of the absolute value of the difference between the heading angles of the two agents, $|\phi_{ij}|$ (see the graphical definition in Fig.~\ref{fig:quantities}). The mean heading difference observed in fish experiments is 27$^\circ$, with a rather narrow PDF associated with a SD of 30$^\circ$, confirming the good level of alignment between the two fish. The DLI simulated pairs are not as aligned as fish pairs, with a larger mean and SD equal to 38$^\circ$, although the Hellinger distance between the two PDF $(H=0.14)$ remains satisfactory. The corresponding PDF for biohybrid pairs exhibits the largest disagreement with the fish experiments of all the PDF presented here ($H=0.30$). Indeed, despite also being peaked at $|\phi_{ij}|=0$, the PDF has a non-negligible weight for $|\phi_{ij}|>90^\circ$, resulting in a much larger mean of $59^\circ$ and a SD of $48^\circ$. This wider PDF is a consequence of the fact that the fish and the LureBot, despite remaining close to each other on average, have a much higher probability than fish pairs to be at a distance above the range of the alignment interaction. Moreover, when the fish and the LureBot are far apart and attempt to get closer, they have a high chance to be actually anti-aligned during this process, hence the significant weight of the PDF near $|\phi_{ij}|=180^\circ$.

Finally, Fig.~\ref{fig:collective}c shows the PDF of the angle of perception $\psi_{ij}$, defined in Fig.~\ref{fig:quantities}. For pairs of fish, the PDF presents clear peaks at $\psi_{ij}=0^\circ$ and near $|\psi_{ij}|=180^\circ$. This indicates that the well aligned fish are following each other rather than swimming side by side. For DLI simulated pairs, the same pattern is observed but with slightly less pronounced peaks, although the Hellinger distance of $H = 0.05$ confirms the excellent agreement between both PDF. As for the biohybrid pair, the PDF averaged over the fish and the LureBot again presents the same peaks as before, but even less pronounced. Again, the less sharp peaks are a consequence of the fact that the biohybrid pairs stand farther from the wall than fish pairs, and above all, of the fact that their distance has a higher probability to be large enough so that their angle of perception $\psi_{ij}$ becomes uncorrelated. The lesser alignment of the biohybrid pairs (see above) originates from the same causes, and in turn also results in a more homogeneous distribution of the angle of perception.
However, the apparent reasonable agreement with the PDF for the fish-only and DLI-SP pairs masks the difference between the PDF for the fish and for the LureBot shown in the top inset of Fig.~\ref{fig:collective}c. There, we observe that the peak near $\psi_{ij}=0^\circ$ is dominated by the contribution of the fish, showing that the fish more often follows the LureBot than the converse. In addition, we find that the PDF for the fish is also peaked slightly above $\psi_{ij}=-180^\circ$, while the PDF for the LureBot has a corresponding peak slightly below $\psi_{ij}=+180^\circ$. By periodicity of 360$^\circ$, these two peaks are obviously located at almost the same angle, but this slight angular shift translates to the fact that the fish is, on average, slightly closer to the wall than the LureBot, as noted in Section~\ref{sec:indquant}.

The instantaneous collective quantities demonstrate that despite the dissimilarities measured in the individual behavior of both DLI-SP and DLI-BP with respect to the fish-only experiment, the collective dynamics are fairly reproduced. Furthermore, the DLI is transferred in a physical system with good agreement compared to its simulated version, and the living agent responds positively. However, the angular control of the robot is arguably less precise, which contributes to the general deviation from the experimental angle-related distributions.

\begin{figure}[ht!]
    \centering
    \includegraphics[width=\linewidth]{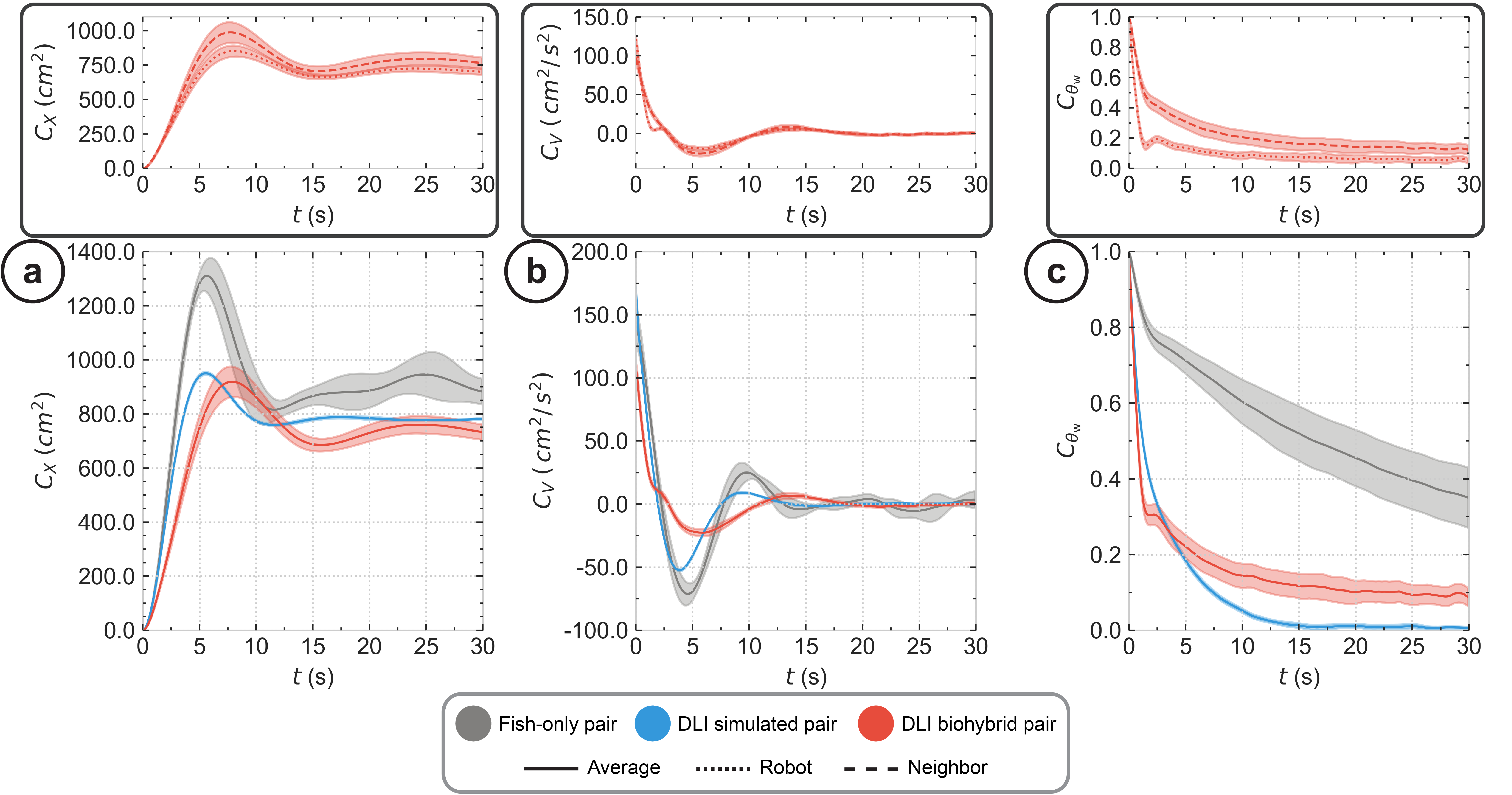}
    \caption{\small
        \textbf{Temporal correlation quantities.} \textbf{(a)} Mean squared displacement $C_{X}(t)$. \textbf{(b)} Velocity autocorrelation $C_V(t)$. \textbf{(c)} Temporal correlations of the angle of incidence to the wall $C_{\theta_{\rm w}}(t)$. Dark gray, blue and red colors correspond to the distributions of the experiment, DLI simulated pairs and DLI biohybrid pairs, respectively. Dotted, dashed and solid lines indicate the robot, neighbor and average agent distributions, respectively. The shaded areas correspond to the standard deviation.
    }
    \label{fig:correlation}
\end{figure}

\subsection{Temporal correlation functions}\label{sec:tempcor}

In Fig.~\ref{fig:correlation}, we plot the three observables used to quantify the temporal correlations that emerge in the system during the long-term dynamics, which are defined in Section~\ref{sec:metrics}.

Fig.~\ref{fig:correlation}a shows the mean square displacement of the agents, $C_X(t)$, in the three considered cases. After a rapid growth, $C_X(t)$ presents a peak and an ultimate decay to a mean level equal to twice the mean square of the distance to the center of the tank. Indeed, for large time difference, the positions at time $t'$ and $t+t'$ become uncorrelated, and we obtain

\begin{eqnarray}
    C_X(t) &=& \left\langle\left[\vec{u}_i(t + t') - \vec{u}_i(t')\right]^ 2\right\rangle\nonumber\\
    &\mathop  \approx \limits_{t \to  + \infty } &\left\langle\vec{u}_i^2(t + t') \right\rangle+ \left\langle\vec{u}_i^2( t') \right\rangle \nonumber\\
    &=& 2\left\langle\vec{u}_i^2( t') \right\rangle,
\end{eqnarray}
which becomes time-independent due to the stationarity of the dynamics. Although $C_X(t)$ has the same qualitative form in the three cases, one observes differences in the position and height of the peak and in the asymptotic value. The latter is explained by the fact that the closer the agents are to the wall, the larger is the mean square of their distance to the center of the tank, $\left\langle\vec{u}_i^2( t') \right\rangle$. Indeed, we have found, in Section~\ref{sec:indquant}, that fish pairs swim closest to the wall, while biohybrid pairs are the farthest, which is consistent with the asymptotic behavior of $C_X(t)$ observed in Fig.~\ref{fig:correlation}a. Furthermore, the top inset of Fig.~\ref{fig:correlation}a for the biohybrid pairs shows that $C_X(t)$ for the fish is systematically larger than for the LureBot, which is also consistent with the fact that the fish swims slightly closer to the wall than the LureBot. As for the position of the peaks in Fig.~\ref{fig:correlation}a, it roughly corresponds to the time for the corresponding agent to travel half of the tank perimeter. This time is directly correlated with the mean speed of the agent. In Section~\ref{sec:indquant}, we found that the fish pairs and DLI simulated pairs had essentially the same mean speed, which explains the agreement between the position of the corresponding peaks in $C_X(t)$. However, we also found that the biohybrid pairs were $20$\,\% slower, which explains the fact that the peak in their $C_X(t)$ is reached at a later time than for fish and DLI pairs.

Fig.~\ref{fig:correlation}b shows the velocity autocorrelation, $C_V(t)$, in the three considered cases, which vanishes for $t$ large enough, when the velocity at time $t+t'$ becomes uncorrelated with that at time $t'$. It can be formally shown that $C_V(t) = \frac{1}{2}\frac{d^2C_X}{dt^2}(t)$ (although this relation is only approximate, when the 2 quantities are observed independently over a finite sampling time), so that the interpretation of the shape of $C_V(t)$ results from the analysis that we have presented above for $C_X(t)$. In particular, the peaks of the first two oscillations in $C_V(t)$ roughly correspond to the two inflection points just before and after the main peak in $C_X(t)$. In addition, $C_V(t=0)$ is the mean square velocity, and we indeed observe an agreement between its value for fish and DLI pairs, while the slower biohybrid pairs result in a lower initial value  of $C_V(t=0)$ in this case.

Finally, the (most subtle) temporal correlation function of the heading  of an agent relative to the wall, $C_{\theta_{\rm w}}(t) = \left\langle\cos\left[\theta_{\rm w}^i(t + t') - \theta_{\rm w}^i(t')\right]\right \rangle$, is shown in Fig.~\ref{fig:correlation}c. For very large time $t$, $C_{\theta_{\rm w}}(t)$ must obviously decay, but we observe that for fish pairs, we  still  have $C_{\theta_{\rm w}}(t = 30\,{\rm s})\approx 0.35$, indicating strong correlations. For DLI simulated pairs, We find that $C_{\theta_{\rm w}}(t)$ vanishes very rapidly ($C_{\theta_{\rm w}}(t = 15\,{\rm s})\approx 0$). Finally, for biohybrid pairs, we still observe some weak remnant correlations at  $t = 30\,{\rm s}$, with $C_{\theta_{\rm w}}(t = 30\,{\rm s})\approx 0.1$ (although the correlation is dominated by the contribution of the fish, as shown in the top inset of Fig.~\ref{fig:correlation}c). Here, the decay rate of $C_{\theta_{\rm w}}(t)$ is strongly related to the sharpness of the peak near $\theta_{\rm w} = 90^\circ$ in the PDF of $\theta_{\rm w}$ (see Fig.~\ref{fig:individual}c and Section~\ref{sec:indquant}). Indeed, a sharp peak suggests that it can take a long time to explore values of $\theta_{\rm w}$ far from $90^\circ$, leading to a slower decay of $C_{\theta_{\rm w}}(t)$. Accordingly, we indeed found that the least sharp peak in the PDF of  $\theta_{\rm w}$ is observed for DLI simulated pairs, resulting in the  fastest decay of $C_{\theta_{\rm w}}(t)$ in this case.

Both the DLI-SP and DLI-BP fail to precisely reproduce the correlation function $C_{\theta_{\rm w}}(t)$, producing a very similar sharp decay compared to the one of real fish. This is again due to the DLI's tendency to frequently produce trajectories farther from the wall than what observed in the experiment. Despite that, the DLI-BP remains fairly faithful to the DLI-SP, which indicates that the DLI is missing some aspects of the social dynamics before being implemented on the robot, but that the robot performs reasonably well in reproducing its underlying model.

\subsection{Complementary results for DLIv2 simulated pairs}

\begin{figure}[ht!]
    \centering
        \includegraphics[width=\linewidth]{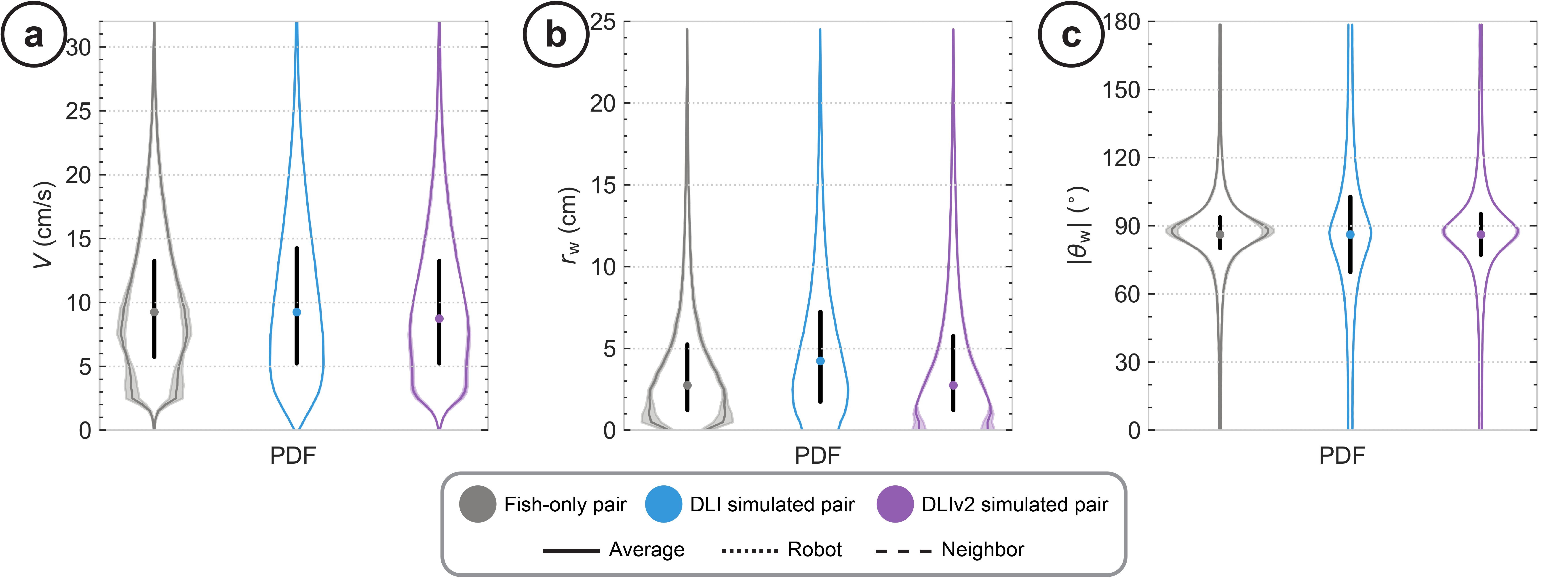}
        \caption{\small
            \textbf{Instantaneous individual quantities.} \textbf{(a)} Speed $V$ probability density function. \textbf{(b)} Distance to the wall $r_{\rm w}$ probability density function. \textbf{(c)} Angle of incidence to the wall $\theta_{\rm w}$ probability density function. Dark gray, blue, and red colors correspond to the distributions of the fish-only experiment, the DLI simulated pairs, and the DLIv2 simulated pairs, respectively. In all PDFs, the colored dot corresponds to the median, and the thick horizontal black line corresponds to the limits of the first and third quartile. The shaded areas correspond to the standard deviation.
        }
        \label{fig:si-individual}
\end{figure}

\begin{figure}[ht!]
    \centering
        \includegraphics[width=\linewidth]{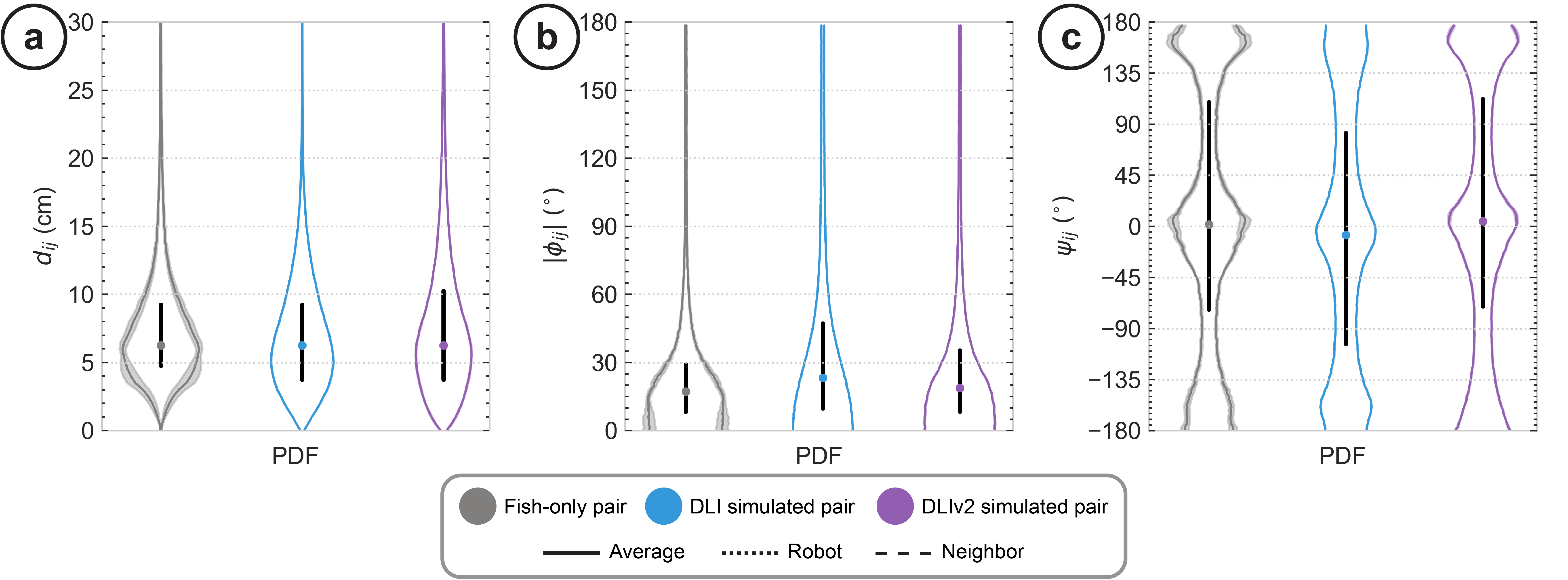}
        \caption{\small
            \textbf{Instantaneous collective quantities.} \textbf{(a)} Interindividual distance $d_{ij}$ probability density function. \textbf{(b)} Difference in heading angles $|\phi_{ij}|$ probability density function. \textbf{(c)} Viewing angle $\psi_{ij}$ probability density function. Dark gray, blue, and red colors correspond to the distributions of the experiment, DLI simulated pairs and DLIv2 simulated pairs, respectively. In all PDFs, the colored dot corresponds to the median, and the thick horizontal black line corresponds to the limits of the first and third quartile. The shaded areas correspond to the standard deviation.
        }
        \label{fig:si-collective}
\end{figure}

\begin{figure}[ht!]
    \centering
        \includegraphics[width=\linewidth]{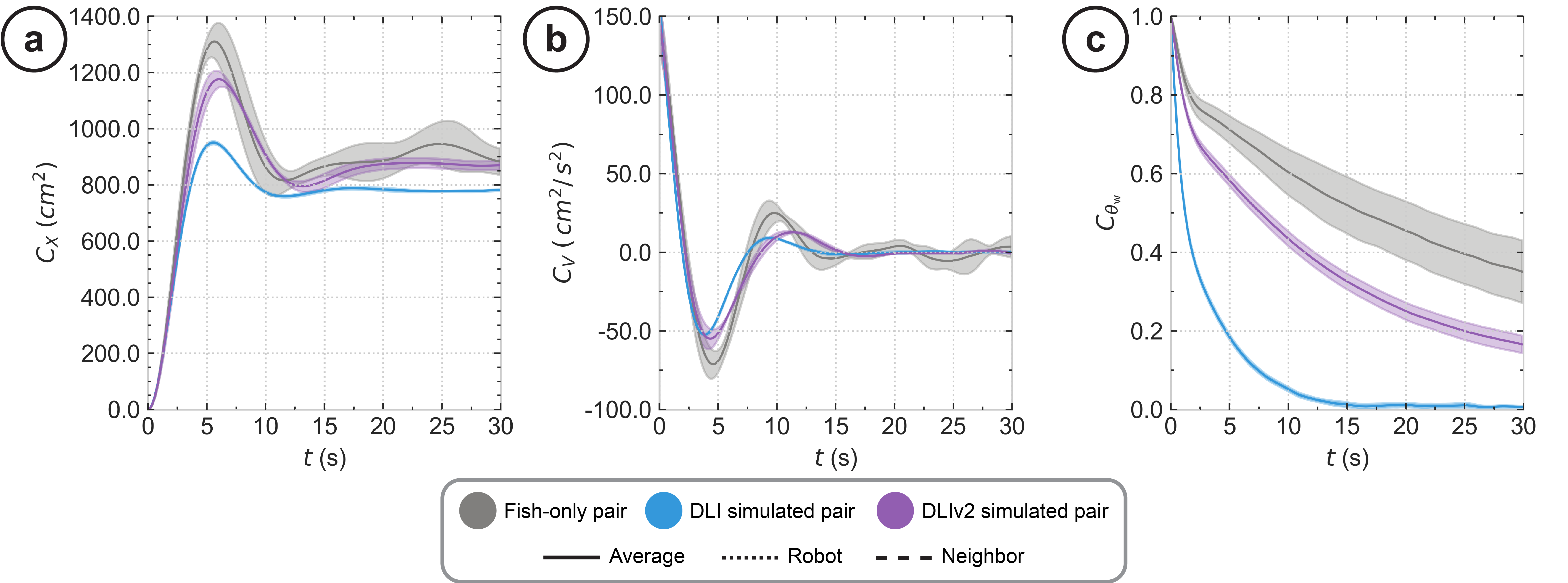}
        \caption{\small
            \textbf{Temporal correlation quantities.} \textbf{(a)} Mean squared displacement $C_{X}(t)$. \textbf{(b)} Velocity autocorrelation $C_V(t)$. \textbf{(c)} Temporal correlations of the angle of incidence to the wall $C_{\theta_{\rm w}}(t)$. Dark gray, blue and red colors correspond to the distributions of the experiment, DLI simulated pairs and DLIv2 simulated pairs, respectively. The shaded areas correspond to the standard deviation.
        }
        \label{fig:si-correlation}
\end{figure}

In addition to the Deep Learning Interaction (DLI) pretrained network utilized in the previous sections, we have also considered an updated version, the DLIv2. This version was retrained on data gathered from the present fish-only experiments under new lighting conditions, concurrently to the robot experiments presented in this work, so that retraining was only feasible after their completion. However, it provided us with the opportunity to test the scalability and predictive performance of the pretrained DLI with new input samples, which, while not fundamentally different, originated from altered social dynamics. For this purpose, we conducted extensive simulations with the DLIv2, and found that their results are in excellent agreement with the present fish-only pair experiments (see Tables~\ref{tab:means_sds} and \ref{tab:hellinger} for further details) for the individual (see Fig.~\ref{fig:si-individual}) and collective (see Fig.~\ref{fig:si-collective}) observables, and for the temporal correlation functions (see Fig.~\ref{fig:si-correlation}). The performance of the simulated DLIv2 model present a significant improvement compared to that of the pretrained DLI model, and one could expect that the LureBot commanded by the DLIv2 model would lead to better results than for the LureBot commanded by the pretrained DLI model. Yet, our point here is that the pretrained DLI model, in different experimental conditions, can still interact with a fish in a similar way as a fish would do.

\section{Discussion and Conclusion}

Despite the wealth of studies on fish-robot interactions, to our knowledge, no prior research has drawn a systematic and quantitative comparison between the social interaction dynamics produced by fish-only, biohybrid, and simulated groups, with a robot  commanded by a machine learning model. This comparison also raises  an intriguing issue: while the reality gap in robotics \cite{mouret201720} typically pertains to the transferability of robot controllers from simulation to real-life conditions, a parallel can be drawn for biohybrid social interactions, termed the biomimicry gap. Addressing this gap is complicated by: 
\begin{itemize}
    \item[(1)] subtle behavioral patterns that behavioral mathematical models or machine learning models fail to capture; 
    \item[(2)] the imperfect or absence of the rendition of the real physics in most models; 
    \item[(3)] the inherently imperfect biomimetic properties of artificial lures and devices.
\end{itemize}
Constructing biohybrid systems with minimal or, ideally, no biomimicry gap, thus making them indistinguishable from pure animal groups, could open doors to groundbreaking research in the study of collective phenomena in animal groups. In particular, this would allow to accurately gauge the reactions of an animal or an animal group to a controlled perturbation (for instance, a robot changing its behavior by adopting a different mean speed or aiming at a target location). Such endeavors require that any non-biomimetic effects of the robot be stringently assessed and resolved. Moreover, it is crucial to ensure that models do not simply overfit experimental data, but can be genuinely transferred to real-world scenarios, through robotic systems that \textit{faithfully} execute the instructions of these models when interacting with animals.

Unfortunately, despite significant strides in the integration of behavioral modeling and robotics hardware, which has long been touted as crucial for deciphering and comprehending the mechanisms underlying collective behavior in animal groups, the closing of the biomimicry gap lacks convincing support in the literature. In particular, several biohybrid implementations tend to limit the set of decisions of the robot (spatial choices, like clockwise/anticlockwise motion) \cite{bonnet2019robots, papaspyros2019bidirectional}. 
In addition, many of these systems rely on simplified passive (open-loop) \cite{phamduy2014fish, polverino2013zebrafish, abaid2012zebrafish, ladu2015acute,  spinello2013acute, ruberto2016zebrafish, bartolini2016zebrafish, kruusmaa2016collective, butail2014influence} or reactive (closed-loop) \cite{cazenille2018mimetic, cazenille2018evolutionary, faria2010novel, swain2011real, landgraf2013interactive, landgraf2016robofish, porfiri2019zebrafish, kim2018closed} models, with only a handful utilizing biomimetic models. Even fewer biomimetic models have been successfully tested in biohybrid groups \cite{cazenille2018mimetic, cazenille2018evolutionary} to emulate real-life dynamics of fish groups. Moreover, to our knowledge, no end-to-end machine learning (ML) model has been examined in this context, despite the booming field of ML. As developed in \cite{papaspyros2024predicting}, assessing a model's fidelity is particularly challenging in the case of ML models, which are often black-box (\textit{i.e.}, not easily explainable). The methodology presented here to validate a model \textit{and} the robotic platform in which it is implemented offers a preliminary solution to this conundrum.
Finally, the importance  of high-fidelity biomimetic lures and agile robotic devices capable of reproducing the typical motion patterns of the considered animal (speed, acceleration...)  is usually underplayed \cite{papaspyros2023biohybrid}. 

In this work, through the precise and comparative quantification of collective behavior in pairs of agents (fish-only pairs, DLI simulated pairs, and DLI biohybrid pairs), we demonstrate that our biomimetic lure and robot system \cite{papaspyros2023biohybrid}, combined with the DLI, are capable of bridging a substantial part of the biomimicry gap. 
More specifically, our study reveals that the overall gap between actual pairs of \textit{H.~rhodostomus} and DLI simulated pairs is fairly small (mean Hellinger distance of $\bar{H}=0.13$), while biohybrid pairs (DLI-BP) and  fish-only pairs are  more dissimilar, but in fair agreement ($\bar{H}=0.20$). Despite this larger difference, the DLI-BP and DLI-SP remain in very good agreement ($\bar{H}=0.11$). We also found that the inherent variability between fish experiments results in an average Hellinger distance of $\bar{H} = 0.10$ (see the end of Table~\ref{tab:hellinger}), which could be considered  as the target performance for future studies following the benchmarking paradigm exploited in this study.
In essence, our DLI model is successful in generating realistic social interactions \cite{papaspyros2024predicting}, our robotic system faithfully replicates its instructions, but the transferred model results in greater discrepancies and the gap widens compared to the simulation (see Table~\ref{tab:hellinger}). Nonetheless, the biohybrid pair is not fully aligned compared to fish groups: the Hellinger distance for the PDF of the angle of incidence to the wall is $H=0.25$ and that of the heading difference is $H=0.30$. These observables are the largest contributors to widening the social interaction discrepancies (\textit{i.e.}, the largest contributor, out of all observables, to increasing the mean Hellinger distance). These discrepancies are consequently observed for the correlation function of the angle of incidence to the wall. However, it is important to recognize that even when comparing two independent fish-only series of experiments, the mean Hellinger distance would not vanish ($\bar{H} = 0.10$, in our case), owing to the inherent variability in fish behavior across experiments. This implies that the main objective for robotic systems should be to notably narrow the gap between experimental results, rather than completely eradicating it.

Moreover, the present work, complementing \cite{papaspyros2023biohybrid,papaspyros2024predicting}, also presents a systematic methodology for a comprehensive assessment of the extent of the biomimicry gap. This is accomplished  by introducing nine observables (easily generalizable to larger groups or other species) that quantify the instantaneous individual and collective behavior, as well as the temporal correlations present in the system. In addition, this methodology is supplemented by the utilization of the Hellinger distance quantifier. We strongly encourage researchers in the field to explore a similar methodology to evaluate the biomimicry gap in their respective systems of study.

Despite the positive results highlighted in our study, we demonstrated that further closing the biomimicry gap necessitates  efforts to minimize all three discrepancy sources depicted in Fig.~\ref{fig:biomimicry_gap}. First, it would require that we refine our modeling approach, \textit{e.g.}, by repeating the biohybrid experiments with the DLIv2 model. Secondly, the physics-related discrepancies, primarily attributable to the transposition of the model into the robot, remains relatively small, but also requires measurable improvement in the robotic system's operation to fully bridge the gap. Finally, discrepancies in the communication cues pose a considerable challenge in terms of evaluation and could only be fully measured in the absence of the other two sources of discrepancies. 

We believe that our study may mark the beginning of many endeavors that integrate animal experiments, biomimetic biohybrid experiments, and simulations of a model commanding the robot, all within a single end-to-end approach. As demonstrated, this approach, combined with a systematic methodology to quantify the biomimicry gap, offers a deeper understanding of the factors contributing to behavioral inaccuracies in biohybrid experiments, thereby highlighting areas in need of improvement. This, in turn, contributes to two main objectives:
\begin{itemize}
    \item[(1)] establishing a more robust experimentation pipeline to explore the diverse sources of the biomimicry gap (such as physical limitations of the robot, social interactions as depicted by the models, and potential discrepancies in the communication cues used to elicit responses);

    \item[(2)] drawing more definitive and insightful behavioral conclusions without the introduction of unrealistic effects inherent in robotic systems and social interaction models.
\end{itemize}

In future research, we aim to extend our experiments to involve multiple individuals and other species, thereby enhancing our understanding of how our robotic platform and DLI model scale to multi-agent interactions and whether large groups of living animals respond similarly to the DLI-driven artificial agent. Additionally, we intend to consistently report the biomimicry gap score, as presented here, with the hope that future studies may adopt a standardized methodology to evaluate the fidelity of biohybrid systems compared to natural ones.

\section{Ethics statement}
\label{sec:ethics}
Experiments were approved by the local ethical committee for experimental animals and were performed at the Centre de Recherches sur la Cognition Animale, Centre de Biologie Int\'egrative, Toulouse, in an approved fish facility (A3155501) under permit APAFIS$\#$27303-2020090219529069~v8 in agreement with the French legislation.

\section{Data accessibility}
The data for the fish-only pairs experiments, DLI pairs numerical simulations, and DLI biohybrid pairs experiments are available at \url{https://doi.org/10.5281/zenodo.8253256}.

\section{Funding}

V.P. and F.M. were supported by the Swiss National Science Foundation project `Self-Adaptive Mixed Societies of Animals and Robots', grant no. 175731. G.T.~and C.S. were supported by the French National Research Agency (ANR-20-CE45-0006-01).

\section{Conflict of interest declaration}

We declare no competing interests.

\section*{References}
\bibliographystyle{amsplain}
\bibliography{bibliography}

\providecommand{\bysame}{\leavevmode\hbox to3em{\hrulefill}\thinspace}
\providecommand{\MR}{\relax\ifhmode\unskip\space\fi MR }
\providecommand{\MRhref}[2]{%
  \href{http://www.ams.org/mathscinet-getitem?mr=#1}{#2}
}
\providecommand{\href}[2]{#2}
\begin{thebibliography}{10}

\bibitem{abaid2012zebrafish}
Nicole Abaid, Tiziana Bartolini, Simone Macr{\`\i}, and Maurizio Porfiri,
  \emph{Zebrafish responds differentially to a robotic fish of varying aspect
  ratio, tail beat frequency, noise, and color}, Behavioural brain research
  \textbf{233} (2012), no.~2, 545--553.

\bibitem{abaid2010fish}
Nicole Abaid and Maurizio Porfiri, \emph{Fish in a ring: spatio-temporal
  pattern formation in one-dimensional animal groups}, Journal of The Royal
  Society Interface \textbf{7} (2010), no.~51, 1441--1453.

\bibitem{bartolini2016zebrafish}
Tiziana Bartolini, Violet Mwaffo, Ashleigh Showler, Simone Macr{\`\i}, Sachit
  Butail, and Maurizio Porfiri, \emph{Zebrafish response to 3d printed shoals
  of conspecifics: the effect of body size}, Bioinspiration \& biomimetics
  \textbf{11} (2016), no.~2, 026003.

\bibitem{basu19972}
Ayanendranath Basu, Ian~R Harris, and Srabashi Basu, \emph{2 minimum distance
  estimation: The approach using density-based distances}, Handbook of
  Statistics \textbf{15} (1997), 21--48.

\bibitem{beran1977minimum}
Rudolf Beran, \emph{Minimum hellinger distance estimates for parametric
  models}, The annals of Statistics (1977), 445--463.

\bibitem{bonnet2014miniature}
Frank Bonnet, Stefan Binder, Marcelo~Elias de~Oliveria, Jos{\'e} Halloy, and
  Francesco Mondada, \emph{A miniature mobile robot developed to be socially
  integrated with species of small fish}, 2014 IEEE International Conference on
  Robotics and Biomimetics (ROBIO 2014), IEEE, 2014, pp.~747--752.

\bibitem{bonnet2018closed}
Frank Bonnet, Alexey Gribovskiy, Jos{\'e} Halloy, and Francesco Mondada,
  \emph{Closed-loop interactions between a shoal of zebrafish and a group of
  robotic fish in a circular corridor}, Swarm Intelligence \textbf{12} (2018),
  227--244.

\bibitem{bonnet2016infiltrating}
Frank Bonnet, Yuta Kato, Jos{\'e} Halloy, and Francesco Mondada,
  \emph{Infiltrating the zebrafish swarm: design, implementation and
  experimental tests of a miniature robotic fish lure for fish--robot
  interaction studies}, Artificial Life and Robotics \textbf{21} (2016), no.~3,
  239--246.

\bibitem{bonnet2019robots}
Frank Bonnet, Rob Mills, Martina Szopek, Sarah Sch{\"o}nwetter-Fuchs, Jos{\'e}
  Halloy, Stjepan Bogdan, Lu{\'\i}s Correia, Francesco Mondada, and Thomas
  Schmickl, \emph{Robots mediating interactions between animals for
  interspecies collective behaviors}, Science Robotics \textbf{4} (2019),
  no.~28, eaau7897.

\bibitem{butail2015fish}
Sachit Butail, Nicole Abaid, Simone Macr{\`\i}, and Maurizio Porfiri,
  \emph{Fish--robot interactions: robot fish in animal behavioral studies},
  Robot Fish: Bio-inspired Fishlike Underwater Robots (2015), 359--377.

\bibitem{butail2014influence}
Sachit Butail, Giovanni Polverino, Paul Phamduy, Fausto Del~Sette, and Maurizio
  Porfiri, \emph{Influence of robotic shoal size, configuration, and activity
  on zebrafish behavior in a free-swimming environment}, Behavioural brain
  research \textbf{275} (2014), 269--280.

\bibitem{calovi2018disentangling}
Daniel~S Calovi, Alexandra Litchinko, Valentin Lecheval, Ugo Lopez,
  Alfonso~P{\'e}rez Escudero, Hugues Chat{\'e}, Cl{\'e}ment Sire, and Guy
  Theraulaz, \emph{Disentangling and modeling interactions in fish with
  burst-and-coast swimming reveal distinct alignment and attraction behaviors},
  PLoS computational biology \textbf{14} (2018), no.~1, e1005933.

\bibitem{cazenille2018evolutionary}
Leo Cazenille, Nicolas Bredeche, and Jos{\'e} Halloy, \emph{Evolutionary
  optimisation of neural network models for fish collective behaviours in mixed
  groups of robots and zebrafish}, Biomimetic and Biohybrid Systems: 7th
  International Conference, Living Machines 2018, Paris, France, July 17--20,
  2018, Proceedings 7, Springer, 2018, pp.~85--96.

\bibitem{cazenille2019automatic}
\bysame, \emph{Automatic calibration of artificial neural networks for
  zebrafish collective behaviours using a quality diversity algorithm}, 2019,
  pp.~38--50.

\bibitem{cazenille2017automated}
Leo Cazenille, Yohann Chemtob, Frank Bonnet, Alexey Gribovskiy, Francesco
  Mondada, Nicolas Bredeche, and Jos{\'e} Halloy, \emph{Automated calibration
  of a biomimetic space-dependent model for zebrafish and robot collective
  behaviour in a structured environment}, 2017, pp.~107--118.

\bibitem{cazenille2018blend}
\bysame, \emph{How to blend a robot within a group of zebrafish: Achieving
  social acceptance through real-time calibration of a multi-level behavioural
  model}, 2018, pp.~73--84.

\bibitem{cazenille2018mimetic}
Leo Cazenille, Bertrand Collignon, Yohann Chemtob, Frank Bonnet, Alexey
  Gribovskiy, Francesco Mondada, Nicolas Bredeche, and Jos{\'e} Halloy,
  \emph{How mimetic should a robotic fish be to socially integrate into
  zebrafish groups?}, Bioinspiration \& biomimetics \textbf{13} (2018), no.~2,
  025001.

\bibitem{chua2018deep}
Kurtland Chua, Roberto Calandra, Rowan McAllister, and Sergey Levine,
  \emph{Deep reinforcement learning in a handful of trials using probabilistic
  dynamics models}, 2018, pp.~4754--4765.

\bibitem{collignon2016stochastic}
Bertrand Collignon, Axel S{\'e}guret, and Jos{\'e} Halloy, \emph{A stochastic
  vision-based model inspired by zebrafish collective behaviour in
  heterogeneous environments}, Royal Society open science \textbf{3} (2016),
  no.~1, 150473.

\bibitem{costa2020automated}
Tiago Costa, Andres Laan, Francisco~JH Heras, and Gonzalo~G de~Polavieja,
  \emph{Automated discovery of local rules for desired collective-level
  behavior through reinforcement learning}, Frontiers in Physics \textbf{8}
  (2020), Article 200.

\bibitem{faria2010novel}
Jolyon~J Faria, John~RG Dyer, Romain~O Cl{\'e}ment, Iain~D Couzin, Natalie
  Holt, Ashley~JW Ward, Dean Waters, and Jens Krause, \emph{A novel method for
  investigating the collective behaviour of fish: introducing ‘robofish’},
  Behavioral Ecology and Sociobiology \textbf{64} (2010), 1211--1218.

\bibitem{harpaz2021collective}
Roy Harpaz, Ariel~C Aspiras, Sydney Chambule, Sierra Tseng, Marie-Ab{\`e}le
  Bind, Florian Engert, Mark~C Fishman, and Armin Bahl, \emph{Collective
  behavior emerges from genetically controlled simple behavioral motifs in
  zebrafish}, Science advances \textbf{7} (2021), no.~41, eabi7460.

\bibitem{heras2018deep}
Francisco J.~H. Heras, Francisco Romero-Ferrero, Robert~C. Hinz, and Gonzalo~G.
  de~Polavieja, \emph{Deep attention networks reveal the rules of collective
  motion in zebrafish}, PLOS Computational Biology \textbf{15} (2019), no.~9,
  1--23.

\bibitem{hochreiter1997long}
Sepp Hochreiter and J{\"u}rgen Schmidhuber, \emph{Long short-term memory},
  Neural computation \textbf{9} (1997), no.~8, 1735--1780.

\bibitem{jayles2020collective}
Bertrand Jayles, Ramon Escobedo, Roberto Pasqua, Christophe Zanon, Adrien
  Blanchet, Matthieu Roy, Gilles Tr{\'e}dan, Guy Theraulaz, and Cl{\'e}ment
  Sire, \emph{Collective information processing in human phase separation},
  Philosophical Transactions of the Royal Society B \textbf{375} (2020),
  no.~1807, 20190801.

\bibitem{kim2018closed}
Changsu Kim, Tommaso Ruberto, Paul Phamduy, and Maurizio Porfiri,
  \emph{Closed-loop control of zebrafish behaviour in three dimensions using a
  robotic stimulus}, Scientific reports \textbf{8} (2018), no.~1, 657.

\bibitem{kruusmaa2016collective}
Maarja Kruusmaa, Guillaume Rieucau, Jos{\'e} Carlos~Castillo Montoya, Riho
  Markna, and Nils~Olav Handegard, \emph{Collective responses of a large
  mackerel school depend on the size and speed of a robotic fish but not on
  tail motion}, Bioinspiration \& biomimetics \textbf{11} (2016), no.~5,
  056020.

\bibitem{ladu2015acute}
Fabrizio Ladu, Violet Mwaffo, Jasmine Li, Simone Macr{\`\i}, and Maurizio
  Porfiri, \emph{Acute caffeine administration affects zebrafish response to a
  robotic stimulus}, Behavioural brain research \textbf{289} (2015), 48--54.

\bibitem{landgraf2016robofish}
Tim Landgraf, David Bierbach, Hai Nguyen, Nadine Muggelberg, Pawel Romanczuk,
  and Jens Krause, \emph{Robofish: increased acceptance of interactive robotic
  fish with realistic eyes and natural motion patterns by live trinidadian
  guppies}, Bioinspiration \& biomimetics \textbf{11} (2016), no.~1, 015001.

\bibitem{landgraf2013interactive}
Tim Landgraf, Hai Nguyen, Stefan Forgo, Jan Schneider, Joseph Schr{\"o}er,
  Christoph Kr{\"u}ger, Henrik Matzke, Romain~O Cl{\'e}ment, Jens Krause, and
  Ra{\'u}l Rojas, \emph{Interactive robotic fish for the analysis of swarm
  behavior}, Advances in Swarm Intelligence: 4th International Conference, ICSI
  2013, Harbin, China, June 12-15, 2013, Proceedings, Part I 4, Springer, 2013,
  pp.~1--10.

\bibitem{miller2012schooling}
Noam Miller and Robert Gerlai, \emph{From schooling to shoaling: patterns of
  collective motion in zebrafish (danio rerio)}, PloS one \textbf{7} (2012),
  no.~11, e48865.

\bibitem{mouret201720}
Jean-Baptiste Mouret and Konstantinos Chatzilygeroudis, \emph{20 years of
  reality gap: a few thoughts about simulators in evolutionary robotics}, 2017,
  pp.~1121--1124.

\bibitem{orger2017zebrafish}
Michael~B Orger and Gonzalo~G de~Polavieja, \emph{Zebrafish behavior:
  opportunities and challenges}, Annual review of neuroscience \textbf{40}
  (2017), 125--147.

\bibitem{papaspyros2019bidirectional}
Vaios Papaspyros, Frank Bonnet, Bertrand Collignon, and Francesco Mondada,
  \emph{Bidirectional interactions facilitate the integration of a robot into a
  shoal of zebrafish danio rerio}, PloS one \textbf{14} (2019), no.~8,
  e0220559.

\bibitem{papaspyros2023biohybrid}
Vaios Papaspyros, Daniel Burnier, Raphaél Cherfan, Guy Theraulaz, Clément
  Sire, and Francesco Mondada, \emph{A biohybrid interaction framework for the
  integration of robots in animal societies}, IEEE Access \textbf{11} (2023),
  67640--67659.

\bibitem{papaspyros2024predicting}
Vaios Papaspyros, Ram{\'o}n Escobedo, Alexandre Alahi, Guy Theraulaz,
  Cl{\'e}ment Sire, and Francesco Mondada, \emph{Predicting the long-term
  collective behaviour of fish pairs with deep learning}, Journal of the Royal
  Society Interface \textbf{21} (2024), no.~212, 20230630.

\bibitem{phamduy2014fish}
P~Phamduy, G~Polverino, RC~Fuller, and M~Porfiri, \emph{Fish and robot dancing
  together: bluefin killifish females respond differently to the courtship of a
  robot with varying color morphs}, Bioinspiration \& biomimetics \textbf{9}
  (2014), no.~3, 036021.

\bibitem{polverino2013zebrafish}
Giovanni Polverino and Maurizio Porfiri, \emph{Zebrafish (danio rerio)
  behavioural response to bioinspired robotic fish and mosquitofish (gambusia
  affinis)}, Bioinspiration \& biomimetics \textbf{8} (2013), no.~4, 044001.

\bibitem{porfiri2018inferring}
Maurizio Porfiri, \emph{Inferring causal relationships in zebrafish-robot
  interactions through transfer entropy: a small lure to catch a big fish},
  Animal Behavior and Cognition \textbf{5} (2018), no.~4, 341--367.

\bibitem{porfiri2019zebrafish}
Maurizio Porfiri, Chiara Spinello, Yanpeng Yang, and Simone Macr{\`\i},
  \emph{Zebrafish adjust their behavior in response to an interactive robotic
  predator}, Frontiers in Robotics and AI \textbf{6} (2019), 38.

\bibitem{romano2019review}
Donato Romano, Elisa Donati, Giovanni Benelli, and Cesare Stefanini, \emph{A
  review on animal--robot interaction: from bio-hybrid organisms to mixed
  societies}, Biological cybernetics \textbf{113} (2019), no.~3, 201--225.

\bibitem{romano2021unveiling}
Donato Romano and Cesare Stefanini, \emph{Unveiling social distancing
  mechanisms via a fish-robot hybrid interaction}, Biological Cybernetics
  \textbf{115} (2021), no.~6, 565--573.

\bibitem{romano2022any}
\bysame, \emph{Any colour you like: fish interacting with bioinspired robots
  unravel mechanisms promoting mixed phenotype aggregations}, Bioinspiration \&
  Biomimetics \textbf{17} (2022), no.~4, 045004.

\bibitem{romano2022robot}
\bysame, \emph{Robot-fish interaction helps to trigger social buffering in neon
  tetras: The potential role of social robotics in treating anxiety},
  International Journal of Social Robotics \textbf{14} (2022), no.~4, 963--972.

\bibitem{ruberto2016zebrafish}
Tommaso Ruberto, Violet Mwaffo, Sukhgewanpreet Singh, Daniele Neri, and
  Maurizio Porfiri, \emph{Zebrafish response to a robotic replica in three
  dimensions}, Royal Society open science \textbf{3} (2016), no.~10, 160505.

\bibitem{spinello2013acute}
Chiara Spinello, Simone Macr{\`\i}, and Maurizio Porfiri, \emph{Acute ethanol
  administration affects zebrafish preference for a biologically inspired
  robot}, Alcohol \textbf{47} (2013), no.~5, 391--398.

\bibitem{swain2011real}
Daniel~T Swain, Iain~D Couzin, and Naomi~Ehrich Leonard, \emph{Real-time
  feedback-controlled robotic fish for behavioral experiments with fish
  schools}, Proceedings of the IEEE \textbf{100} (2011), no.~1, 150--163.

\bibitem{zienkiewicz2018data}
Adam~K Zienkiewicz, Fabrizio Ladu, David~AW Barton, Maurizio Porfiri, and Mario
  Di~Bernardo, \emph{Data-driven modelling of social forces and collective
  behaviour in zebrafish}, Journal of Theoretical Biology \textbf{443} (2018),
  39--51.

\end{thebibliography}

\end{document}